\def\year{2022}\relax
\documentclass[letterpaper]{article} 
\usepackage{aaai22}  
\usepackage{times}  
\usepackage{helvet}  
\usepackage{courier}  
\usepackage[hyphens]{url}  
\usepackage{graphicx} 
\usepackage{natbib}  
\usepackage{caption} 
\DeclareCaptionStyle{ruled}{labelfont=normalfont,labelsep=colon,strut=off} 
\usepackage{algorithm}
\usepackage{algorithmic}
\usepackage[export]{adjustbox}

\usepackage{multicol}
\usepackage{graphicx}
\usepackage{amsfonts}
\usepackage{amsmath}
\usepackage{placeins}
\usepackage{color, soul}
\usepackage[labelformat=empty, position=top]{subcaption}
\usepackage{bbm}

\usepackage[algo2e, ruled, vlined]{algorithm2e}
\usepackage{amsthm}
\usepackage{tikz}

\newtheorem{proposition}{Proposition}
\newtheorem*{proposition*}{Proposition}

\theoremstyle{definition}
\newtheorem{problem}{Problem}

\theoremstyle{remark}
\newtheorem{remark}{Remark}
\newtheorem*{remark*}{Remark}
\newtheorem{example}{Example}
\DeclareMathOperator*{\argmax}{arg\,max}
\DeclareMathOperator*{\argmin}{arg\,min}
\DeclareMathOperator*{\cumul}{\textrm{cumul}_\mathit{C}}
\DeclareMathOperator*{\fin}{\textrm{fin}}
\DeclareMathOperator*{\inff}{\textrm{inf}}

\usepackage{newfloat}
\usepackage{listings}
\floatstyle{ruled}
\newfloat{listing}{tb}{lst}{}
\floatname{listing}{Listing}
\title{Planning for Risk-Aversion and Expected Value in MDPs}

\author{
	Marc Rigter, 
	Paul Duckworth, 
	Bruno Lacerda, 
	Nick Hawes
}
\affiliations{
	Oxford Robotics Institute, University of Oxford, United Kingdom \\
	\{mrigter, pduckworth, bruno, nickh \}@robots.ox.ac.uk
}

\usepackage{bibentry}

\begin{document}
	
\maketitle

\begin{abstract}
Planning in Markov decision processes (MDPs) typically optimises the \emph{expected} cost.
However, optimising the expectation does not consider the risk that for any given run of the MDP, the total cost received may be unacceptably high.
An alternative approach is to find a policy which optimises a risk-averse objective such as conditional value at risk (CVaR).
However, optimising the CVaR alone may result in poor performance in expectation.
In this work, we begin by showing that there can be multiple policies which obtain the optimal CVaR.
This motivates us to propose a lexicographic approach which minimises the expected cost subject to the constraint that the CVaR of the total cost is optimal. 
We present an algorithm for this problem and evaluate our approach on four domains.
Our results demonstrate that our lexicographic approach improves the expected cost compared to the state of the art algorithm, while achieving the optimal CVaR.
\end{abstract}

\section{Introduction}

\begin{figure}[t!]
	\centering
	\includegraphics[width=0.8\columnwidth]{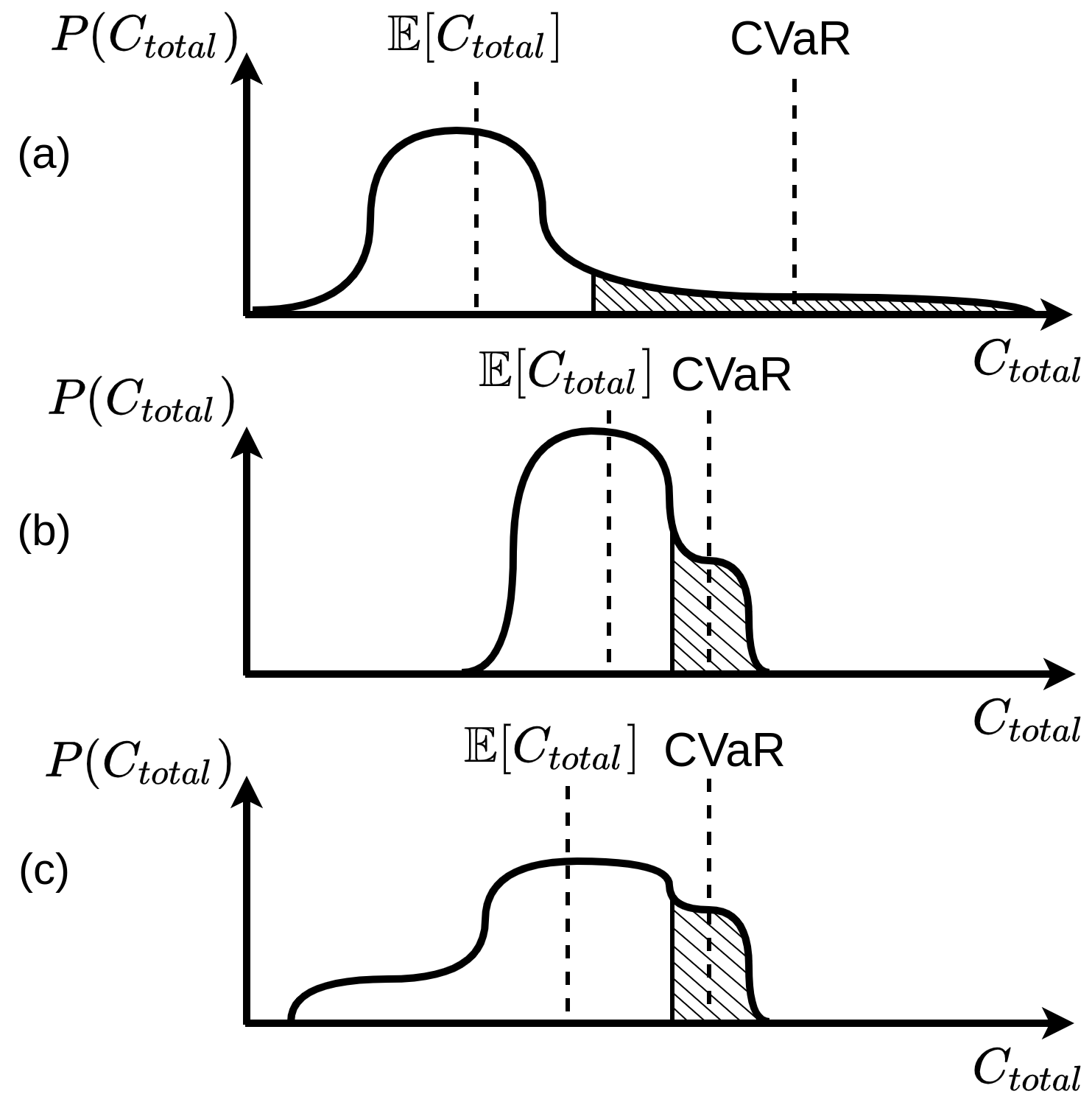}
	\caption{Illustration of distributions over the total cost, $C_{total}$, for different approaches: (a) expected value optimisation, (b) CVaR optimisation, and (c) our approach, which optimises the expected value subject to the constraint that CVaR is optimal. Shaded regions indicate the $\alpha$-portion of the right tail of each distribution. \label{fig:cvar_drawing}}
	\vspace{-3mm}
\end{figure}

Markov decision processes (MDPs) are a common framework for decision-making under uncertainty, and have been applied to many domains such as inventory control~\cite{ahiska2013markov} and  robot navigation~\cite{lacerda2019probabilistic}. 
The solution for an MDP typically optimises the expected total return, defined as either a reward to maximise or a cost to minimise. In this work, we consider the cost minimisation setting (Fig~\ref{fig:cvar_drawing}a). For any single run of the MDP, the total cost received is uncertain due to the MDP's stochastic transitions. In some applications we wish to compute \emph{risk-averse} policies which prioritise avoiding the worst outcomes, rather than simply minimising the expected total cost irrespective of the variability.

As an example, consider an inventory control problem where a decision-maker decides how much stock to purchase each day, while subject to uncertain demand from customers. 
The policy which optimises expected value purchases large quantities of stock to maximise the expected profit.
However, if demand is much lower than expected, significant losses may be incurred. 
A risk-averse policy purchases less stock, and is therefore less profitable on average, but avoids the possibility of large losses.

In this work, we focus on \textit{conditional value at risk} (CVaR), a well known coherent risk metric~\cite{rockafellar2000optimization}, in the \textit{static} risk setting. In MDPs, the static CVaR at confidence level $\alpha \in (0, 1]$ corresponds to the mean total cost in the worst $\alpha$-fraction of runs. 
Therefore, optimising the CVaR corresponds to optimising the $\alpha$-portion of the right tail of the distribution over the total cost (Fig.~\ref{fig:cvar_drawing}b).
%

%
%

%

\setlength{\textfloatsep}{18pt}

We propose that for risk-sensitive applications the~\textit{primary} objective should be to avoid the risk of a poor outcome (i.e. avoid significant losses).  
However, if the risk of a poor outcome has been avoided, then the~\textit{secondary} objective should be to optimise the expected value (i.e. maximise the expected profit).
This motivates us to propose a lexicographic approach that optimises the expected total cost subject to the constraint that the CVaR of the total cost is optimal. 
%
During execution, the resulting policy is initially risk-averse.
However, the policy may begin taking more aggressive actions to improve the expected cost, provided that there is no longer any risk of incurring a bad run which would influence the CVaR.
%
%
Fig.~\ref{fig:cvar_drawing}c illustrates the cost distribution for our approach, compared to optimising the expected cost or the CVaR only. 
As indicated by the dashed lines, our approach obtains the same CVaR as optimising for CVaR alone (Fig.~\ref{fig:cvar_drawing}b), but the expected cost is improved.

%
Our main contributions are: 1) showing that there can be multiple policies that obtain the optimal CVaR in an MDP, 2) proposing and formalising the lexicographic problem of optimising expected value in MDPs subject to the constraint of minimising CVaR, and 3) an algorithm to solve this problem, based on reducing it to a two-stage optimisation in a stochastic game.
\noindent To the best of our knowledge, this is the first work to address lexicographic optimisation of CVaR and expected value in sequential decision making problems. 

We evaluate our algorithm on four domains, including a road network navigation domain which uses real data from traffic sensors~\cite{chen2001freeway} to simulate journey times. Our experimental results demonstrate that our approach significantly improves the expected cost on all domains while attaining the optimal conditional value at risk.

\vspace{-0.5mm}
\section{Related Work}

Many existing works address risk-averse optimisation in MDPs. Early work considered the expected utility framework~\cite{howard1972risk}, where the cost is transformed according to a convex utility function to achieve risk-aversion. However, it is difficult to ``shape'' the cost function appropriately to achieve the desired behaviour~\cite{majumdar2020should}. Other works consider risk metrics such as the mean-variance criterion~\cite{sobel1982variance}, or the value at risk~\cite{filar1995percentile}. However, these risk metrics are not \textit{coherent}, meaning that they do not satisfy properties consistent with rational decision-making. 
%
%
 See~\citet{artzner1999coherent} or~\citet{majumdar2020should} for an introduction to coherent risk metrics.

Examples of coherent risk metrics include the conditional value at risk (CVaR)~\cite{rockafellar2000optimization}, the entropic value at risk~\cite{ahmadi2012entropic}, and the Wang risk measure~\cite{wang2000class}. In this work we focus on CVaR because it is intuitive to understand, and it is the prevailing risk metric used in risk-sensitive domains such as finance~\cite{basel2014fundamental}.
We consider \textit{static} risk, where the risk metric is applied to the total cost, rather than \textit{dynamic} risk which penalises risk at each time step~\cite{ruszczynski2010risk}. For the static CVaR setting, approaches based on value iteration over an augmented state space have been proposed~\cite{bauerle2011markov, chow2015risk}. Other works propose policy gradient methods to find a locally optimal solution for CVaR~\cite{borkar2010risk, chow2014algorithms, tamar2014optimizing, tamar2015policy, prashanth2014policy, chow2017risk, tang2019worst}, or general coherent risk metrics~\cite{tamar2016sequential}.~\citet{keramati2020being} present an RL algorithm based on optimism in the face of uncertainty, while~\citet{rigter2021risk} propose an approach based on Monte Carlo tree search and also consider model uncertainty.

Existing works have considered trading off the conflicting objectives of expected cost and risk.~\citet{petrik2012an} optimise a weighted combination of dynamic CVaR and expected value. For any given set of weightings, the performance for conditional value at risk may be sacrificed to improve the expected value. In contrast, our lexicographic approach ensures that the optimal CVaR is always attained.


An alternative approach to finding risk-averse solutions to MDPs is to impose constraints. Constrained MDPs optimise the expected value subject to a constraint on the expected value of a secondary cost function~\cite{altman1999constrained, santana2016rao}. By adding cost penalties to undesirable states, constraints may be used to discourage unwanted behaviour. However, this approach only constrains the expected cost and it is unclear how to choose cost penalties which result in the desired risk-averse behaviour. 


CVaR-constrained problems have been addressed by existing works~\cite{chow2014algorithms, chow2017risk, borkar2010risk, prashanth2014policy}. In this approach, the user defines a CVaR threshold. The expected value is optimised subject to a constraint that the CVaR is less than the chosen threshold. A disadvantage of this approach is that it may be difficult to choose an appropriate CVaR threshold which is both feasible, and results in the desired level of risk-aversion.

Lexicographic approaches to multi-objective decision making in MDPs have been proposed~\cite{mouaddib2004multi, wray2015multi, lacerda_sdmia15}. These approaches optimise the expected value for each objective in a lexicographic ordering. Unlike these approaches, we consider risk-averse planning. To the best of our knowledge, this is the first work to propose a lexicographic approach to the optimisation of CVaR and expected value in sequential decision making problems.

\vspace{-0.5mm}
\section{Preliminaries}
\subsection{Conditional Value at Risk}
\label{sec:prelim_cvar}
Let $Z$ be a bounded-mean random variable, i.e. $E[|Z|] < \infty$, on a probability space $(\Omega, \mathcal{F}, \mathcal{P})$, with cumulative distribution function ${F(z) = \mathcal{P}(Z \leq z)}$. In this paper, we interpret $Z$ as the total cost which is to be minimised. The \textit{value at risk} (VaR) at confidence level $\alpha \in (0, 1]$ is defined as VaR$_\alpha (Z) = \min \{z | F(z) \geq 1 - \alpha\}$. The \textit{conditional value at risk} at confidence level $\alpha$ is defined as
\begin{equation}
	\label{eq:cvar_def}
	\textnormal{CVaR}_\alpha (Z) = \frac{1}{\alpha} \int_{1-\alpha}^1 \textnormal{VaR}_{1-\gamma}(Z) d\gamma.
\end{equation}

If $Z$ has a continuous distribution, $\textnormal{CVaR}_\alpha (Z)$ can be defined using the more intuitive expression: ${\textnormal{CVaR}_\alpha (Z) = \mathbb{E}[Z \mid  Z \geq \textnormal{VaR}_\alpha (Z)] }$. Thus, $\textnormal{CVaR}_\alpha(Z)$ may be interpreted as the expected value of the $\alpha$-portion of the right tail of the distribution of $Z$.
CVaR may also be defined as the expected value under a perturbed distribution using its dual representation~\cite{rockafellar2000optimization, shapiro2014lectures}:
\begin{equation}
	\label{eq:repr_theorem}
	\textnormal{CVaR}_\alpha (Z) = \max_{\xi \in \mathcal{B}(\alpha, \mathcal{P})} \mathbb{E}_\xi \big[Z \big],
\end{equation}

\noindent where $\mathbb{E}_\xi [Z]$ denotes the $\xi$-weighted expectation of $Z$, and the \textit{risk envelope}, $\mathcal{B}$, is given by
\begin{equation}
	\label{eq:risk_envelope}
	\mathcal{B}(\alpha, \mathcal{P}) = \Bigg\{ \xi \ \mid \ \xi(\omega) \in \Big[0, \frac{1}{\alpha} \Big], \int_{\omega \in \Omega} \xi(\omega) \mathcal{P}(\omega) d\omega = 1 \Bigg\}.
\end{equation}

\noindent where $\mathcal{P}(\omega)$ is the probability density function if $Z$ is continuous, and the probability mass function if $Z$ is discrete.

Therefore, the CVaR of a random variable $Z$ may be interpreted as the expectation of $Z$ under a worst-case perturbed distribution, $\xi \mathcal{P}$. The risk envelope is defined so that the probability of any outcome can be increased by a factor of at most $1/\alpha$, whilst ensuring the perturbed distribution is a valid probability distribution.

\subsection{Markov Decision Processes}
\label{sec:mdps}
In this work, we consider stochastic shortest path (SSP) Markov decision processes (MDPs). An SSP MDP is a tuple, $\mathcal{M} = (S, A, C, T, G, s_0)$, where $S$ and $A$ are finite state and action spaces; $C : S \times A \rightarrow \mathbb{R}_+$ is the cost function; $T : S \times A \times S \rightarrow [0, 1]$ is the probabilistic transition function; $G \subset S$ is the set of absorbing  goal states, from which the model incurs zero cost; and $s_0$ is the initial state. 
A history of $\mathcal{M}$ is a sequence $h = s_0 a_0 s_1 a_1\ldots $ such that   $T(s_i,a_i,s_{i+1})>0$ for all $i \in \{0,\ldots,|h|\}$, where $|h|$ denotes the length of $h$.
We denote the set of all finite-length histories over $\mathcal{M}$ as   $\mathcal{H}^{\mathcal{M}}_{\fin}$ and  the set of all infinite-length histories over $\mathcal{M}$ as   $\mathcal{H}^{\mathcal{M}}_{\inff}$, and define the set of all histories over $\mathcal{M}$ as $\mathcal{H}^{\mathcal{M}}=\mathcal{H}^{\mathcal{M}}_{\fin} \cup \mathcal{H}^{\mathcal{M}}_{\inff}$.
The cumulative cost  function $\cumul:\mathcal{H}^\mathcal{M} \rightarrow \mathbb{R}^+$ is defined such that, given history $h=s_0a_0 s_1 a_1 \ldots $,  $\cumul(h) = \sum_{t=0}^{|h|} C(s_t,a_t)$.
A history-dependent policy is a function $\pi:\mathcal{H}^{\mathcal{M}}_{\fin} \rightarrow A$, and we write $\Pi^\mathcal{M}_\mathcal{H}$ to denote the set of all history-dependent policies for $\mathcal{M}$. 
If $\pi$ only depends on the last state $s_t$ of $h$, then we say $\pi$ is \emph{Markovian}, and we denote the set of all Markovian policies as $\Pi^\mathcal{M}$.
A policy $\pi$ induces a probability distribution $\mathcal{P}^\mathcal{M}_\pi$ over $\mathcal{H}^{\mathcal{M}}_{\inff}$, and we define  the cumulative cost  distribution $\mathcal{C}_\pi^\mathcal{M}$  as the distribution over the value of $\cumul$ for infinite-length histories of $\mathcal{M}$ under policy $\pi$.
%
%
A policy is proper at $s$ if it reaches $s_g \in G$ from $s$ with probability~1.
A policy is proper if it is proper at all states. In an SSP MDP, the following assumptions are made~\cite{kolobov2012planning}: a) there exists a proper policy, and b) every improper
policy incurs infinite cost at all states where it is improper.
These assumptions ensure the expected value of the cumulative cost distribution is finite for at least one policy in the SSP. \looseness=-1

\subsection{Stochastic Games}
In this paper, we formulate CVaR optimisation as a turn-based two-player zero-sum stochastic shortest path game (SSPG). An SSPG between an agent and an adversary is a generalisation of an SSP MDP and can be defined using a similar tuple  $\mathcal{G}=(S, A, C, T, G, s_{0})$. 
The elements of $\mathcal{G}$ are interpreted as with MDPs, but are extended to model a two-player game.
$S$ is partitioned into a set of agent states $S_{agt}$, and a set of adversary states $S_{adv}$.
Similarly,  $A$ is partitioned into a set of agent actions $A_{agt}$, and a set of adversary actions $A_{adv}$.
The transition function is defined such that agent actions can only be executed in agent states, and adversary actions can only be executed in adversary states.

We denote the set of Markovian agent policies mapping agent states to agent actions as $\Pi^\mathcal{G}$ and the set of Markovian adversary policies, defined similarly, as $\Sigma^\mathcal{G}$.
Similar to MDPs, a pair $(\pi,\sigma)$ of agent-adversary policies induces a probability distribution $\mathcal{P}^\mathcal{G}_{(\pi,\sigma)}$ over infinite-length histories, and we define  $\mathcal{C}_{(\pi,\sigma)}^\mathcal{G}$ as the cumulative cost distribution of $\mathcal{G}$ under $\pi$ and $\sigma$.
In an SSPG, the agent seeks to minimise the expected cumulative cost, whilst the adversary seeks to maximise it:
\begin{equation} \label{eq:ssgp}
\min_{\pi \in \Pi^\mathcal{G}}\  \max_{\sigma \in \Sigma^\mathcal{G}} \mathbb{E} \big[\mathcal{C}_{(\pi,\sigma)}^\mathcal{G}\big].
\end{equation}

In an SSPG, two assumptions are made to ensure that the value in (\ref{eq:ssgp}) is finite: a) there exists a policy for the agent which is proper
for all possible policies of the adversary, and b) for any states where $\pi$ and $\sigma$ are improper, the expected cost for the agent
is infinite \cite{patek1999stochastic}.

\subsection{CVaR Optimisation in MDPs}
\label{sec:chow_approach}

Many existing works have addressed the problem of optimising the CVaR of $\mathcal{C}_\pi^\mathcal{M}$, defined as follows.

\medskip
\begin{problem} 
	\label{prob:mdp_cvar}
	Let $\mathcal{M}$ be an MDP.
	Find the  optimal CVaR of the cumulative cost at confidence level $\alpha$:
	\begin{equation}
		\label{eq:mdp_cvar}
		\min_{\pi \in \Pi^\mathcal{M}_\mathcal{H}} \textnormal{CVaR}_\alpha (\mathcal{C}_\pi^\mathcal{M}).
	\end{equation} 
\end{problem}

Note that the optimal policy for Problem~\ref{prob:mdp_cvar} may be history-dependent~\cite{bauerle2011markov}.
Methods based on dynamic programming have been proposed to solve Problem~\ref{prob:mdp_cvar} \cite{bauerle2011markov, chow2015risk, pflug2016time}. In particular, the current state of the art approach~\cite{chow2015risk} formulates Problem~\ref{prob:mdp_cvar} as the expected value in an SSPG against an adversary which modifies the transition probabilities. This formulation is based on the CVaR representation theorem in Eq.~\ref{eq:repr_theorem}, where CVaR may be represented as the expected value under a perturbed probability distribution. The SSPG is defined so that the ability for the adversary to perturb the transition probabilities corresponds to the risk envelope given in Eq.~\ref{eq:risk_envelope}.

Formally, the CVaR SSPG is defined by the tuple ${\mathcal{G}^+ = (S^+, A^+, C^+, T^+, G^+, s^+_{0})}$.
The state space $S^+ = S  \times [0, 1] \times (A \cup \{ \bot \})$ is the original MDP state space augmented with a continuous state factor, $y \in [0,1]$, representing the ``budget'' of the adversary to perturb the probabilities; and a state factor $a \in A \cup \{ \bot \}$ indicating the most recent agent action if it is the adversary's turn to choose an action, and $\bot$ if is the agent's turn.
The action space is defined as $A^+ = A \cup \Xi$, i.e. the agent actions are the actions in the original MDP, and the adversary actions are a set $\Xi$ that will be defined next.
%

The CVaR SSPG transition dynamics are as follows. Given an agent state, $(s, y, \bot) \in S^+_{agt}$, the agent applies an action, $a \in A$, and receives cost $C^+((s, y, \bot), a) = C(s, a)$. The state then transitions to the adversary state $(s, y, a) \in S^+_{adv}$. The adversary then chooses an action  to perturb the original MDP transition probabilities from a continuous action space  defined as:
\begin{multline}
\label{eq:adv_actions}
\Xi(s, y, a) = \Big\{\xi \in \mathbb{R}^{|S|}\ \big|\ 0 \leq  \xi(s') \leq \eta \ \ \forall s'   \\ \textnormal{ and } \sum_{s' \in S } \big[ \xi(s') \cdot T(s, a, s') \big] = 1  \Big\},
\end{multline}

\noindent where $\eta = \infty$ if $y=0$, and $\eta = 1/y$ otherwise; $T$ is the original MDP transition function.  Eq.~\ref{eq:adv_actions} restricts the adversary actions so the probability of any history is increased by at most $1/y$, and the perturbed transition probabilities remain a valid probability distribution.
After the adversary chooses the perturbation action $\xi \in \Xi(s, y, a) $, the game transitions back to an agent state $(s', y \cdot \xi(s'), \bot) \in S^+_{agt}$ according to the following transition function where the probability of each successor $s'$ in the original MDP is perturbed by the factor $\xi(s')$:
\begin{equation}
\label{eq:adv_succs}
T^+((s, y, a), \xi, (s', y \xi (s'), \bot)) = 
\xi(s')  T(s, a, s').
\end{equation}

Finally, we define the initial augmented state as  ${s_{0}^+ = (s_0, \alpha, \bot) \in S^+_{agt}}$, where the $y$ state variable is set to the CVaR confidence level $\alpha$. We also define $G^+$ as the set of goal states on the augmented state space corresponding to goal states in the original MDP.

\citet{chow2015risk} showed that the minimax expected value equilibrium for $\mathcal{G}^+$ corresponds to the optimal CVaR.

\begin{proposition}\cite{chow2015risk}
Let $\mathcal{G}^+$ be a CVaR SSPG corresponding to original MDP $\mathcal{M}$. Then:
\begin{equation}
	\label{eq:prop1}
	\min_{\pi \in \Pi_\mathcal{H}^\mathcal{M}} \textnormal{CVaR}_\alpha (\mathcal{C}^\mathcal{M}_\pi) = \min_{\pi \in \Pi^\mathcal{G^+}}\  \max_{\sigma \in \Sigma^\mathcal{G^+}} \mathbb{E} \big[\mathcal{C}_{(\pi,\sigma)}^\mathcal{G^+}\big].
\end{equation}
\label{prop:chow}
\end{proposition}

Proposition~\ref{prop:chow} holds because the $y$ state variable keeps track of the total multiplicative perturbation to the probability of any history. Thus, the constraints in Eq.~\ref{eq:adv_actions} ensure that the maximum perturbation to the probability of any history from the initial state is $1/\alpha$, and that the perturbed distribution over histories is a valid probability distribution. Therefore, the admissible adversarial perturbation actions in $\mathcal{G}^+$ correspond to the risk envelope in Eq.~\ref{eq:risk_envelope}. According to Eq.~\ref{eq:repr_theorem}, the CVaR is the expected value under the perturbations in the risk-envelope that maximise the expected cost.

To compute the solution to Eq.~\ref{eq:prop1}, we denote the value function for the augmented state by $V_{\textnormal{CV}}(s, y, \bot) = \min_{\pi \in \Pi_\mathcal{H}^\mathcal{M}} \textnormal{CVaR}_y (\mathcal{C}^\mathcal{M}_\pi)$. This value function can be computed using minimax value iteration over $\mathcal{G}^+$ using the following Bellman equation:
\vspace{-2mm}
\begin{multline}
	\label{eq:cvar_vi}
	V_{\textnormal{CV}}(s, y, \bot) = \min_{a \in A} \Bigg[C(s, a) +  \\ \max_{\xi \in \Xi(s, y, a)}  \sum_{s' \in S} \xi(s') T(s, a, s') V_{\textnormal{CV}}(s', y\xi(s'), \bot) \Bigg].
\end{multline}

We denote the policy corresponding to the value function obtained by solving Eq.~\ref{eq:cvar_vi} by $\pi_{\textnormal{CV}}$. As the augmented state space contains a continuous variable and the action space is also continuous,~\citet{chow2015risk} use approximate dynamic programming with linear function approximation. 

\setlength{\textfloatsep}{20pt}
\begin{figure}[t!bh]
	\centering
	\begin{subfigure}[t]{0.05\textwidth}
		\vspace{15mm}
		(a)
	\end{subfigure}
	\begin{subfigure}[t]{0.33\textwidth}
		\includegraphics[width=1\linewidth, valign=t]{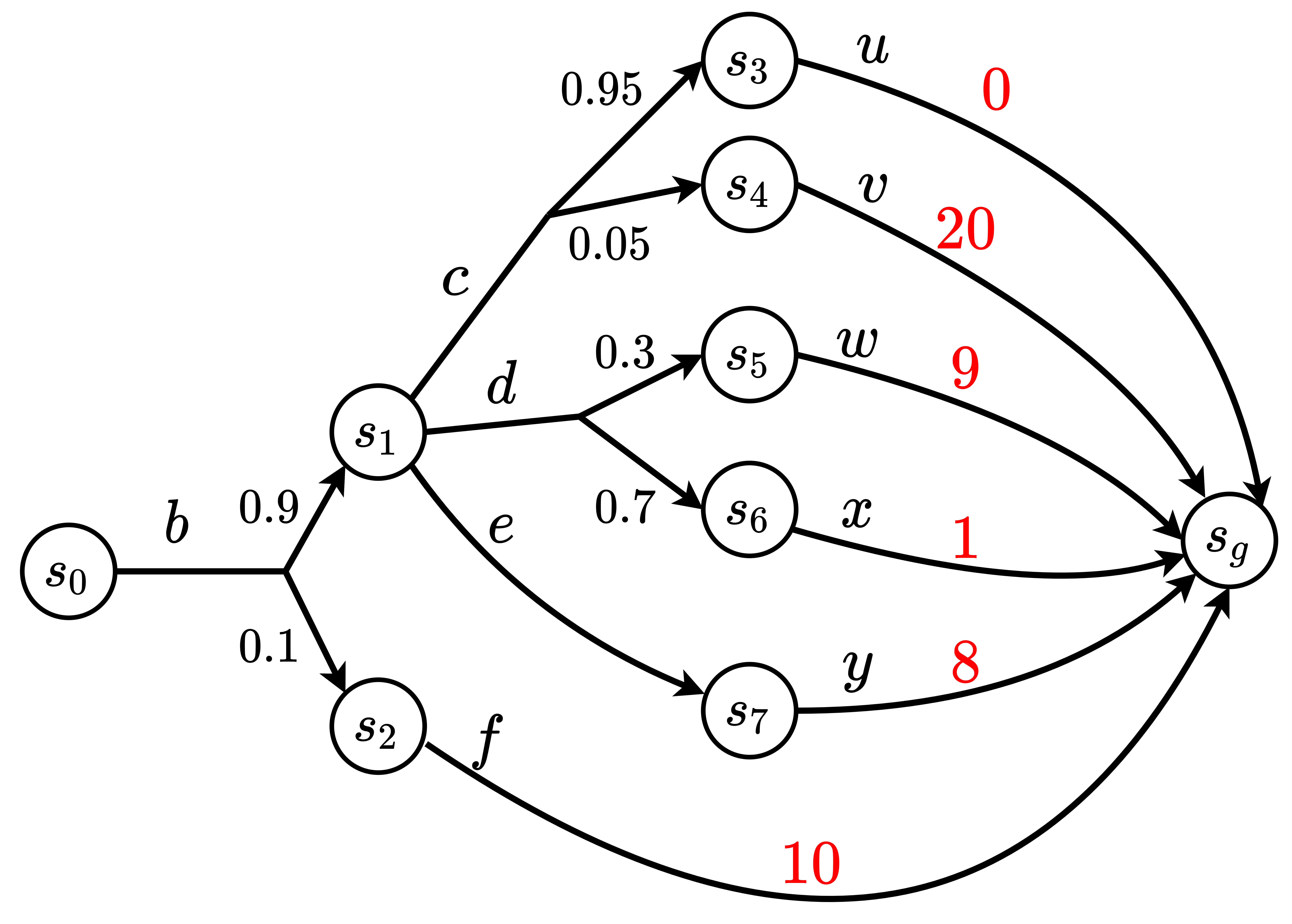}
		\caption{}
		\label{fig:example_mdp}
	\end{subfigure}
	
	\vspace{-3mm}
	\begin{subfigure}[t]{0.03\textwidth}
		\vspace{2mm}
		(b)
	\end{subfigure}
	\begin{subfigure}[t]{0.41\textwidth}
		\includegraphics[width=1\linewidth, valign=t]{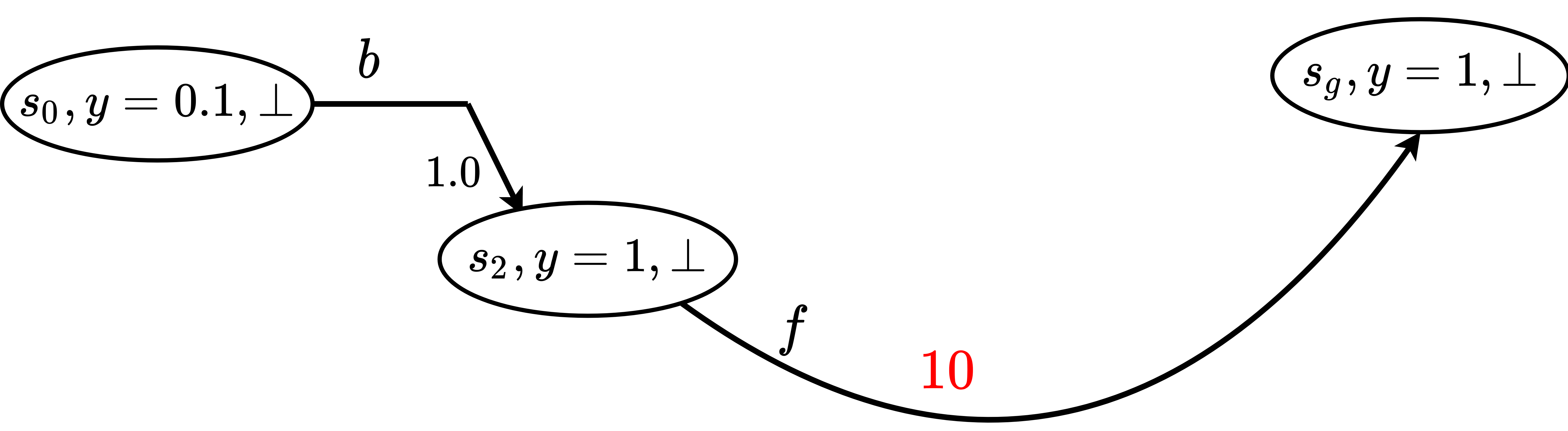}
		\caption{}
		\label{fig:example_mc}
		\vspace{-2mm}
	\end{subfigure}
	\vspace{-2mm}
	\caption{(a) Example Markov decision process with initial state $s_0$ and goal state $s_g$. Letters indicate actions. Costs are assumed to be zero unless indicated otherwise in red. (b) Corresponding Markov chain induced by the solution to Equation~\ref{eq:prop1} when $\alpha = 0.1$.}
	\label{fig:example}
	\vspace{-4mm}
\end{figure}

\begin{example}
	\label{example:1}
 In Fig.~\ref{fig:example} we present an illustrative example of the approach to CVaR optimisation from \citet{chow2015risk}.
Fig.~\ref{fig:example_mdp} depicts an example MDP.
Fig.~\ref{fig:example_mc} shows the  solution to Eq.~\ref{eq:prop1} to optimise the CVaR when $\alpha = 0.1$, illustrated as a Markov chain (MC): 
the adversary perturbs the transition probabilities so that the probability of transitioning from $s_0$ to $s_2$ is increased from $0.1$ to $1$ and, conversely, the probability of transitioning from $s_0$ to $s_1$ is decreased from $0.9$ to $0$.
The expected value in this MC is 10, and by Proposition~\ref{prop:chow} this is the optimal CVaR at $\alpha = 0.1$ in the original MDP.
\end{example}

\begin{remark}
	\label{remark:1}
	State $s_1$ is reachable in the original MDP, but assigned zero probability by the adversary.
	At states assigned zero probability by the adversary, $y=0$ in the CVaR SSPG.
	When $y=0$, the adversary has unlimited power to perturb the transition probabilities (Eq.~\ref{eq:adv_actions}).
	Therefore, when $y=0$ the minimax value iteration in Eq.~\ref{eq:cvar_vi} optimises for the minimum worst-case cost, as the adversary has the power to make the agent transition deterministically to the worst possible successor state for all future transitions. 
	We will refer to the approach from~\citet{chow2015risk} as \textit{CVaR-Worst-Case} because of this property that for histories assigned zero probability by the adversary, this method optimises for the minimum worst-case total cost.
	In the following section we show that this behaviour is unnecessarily conservative to optimise CVaR.
	We propose a method for finding an alternative policy to execute in such situations that optimises the expected value while maintaining the optimal CVaR.
\end{remark}

%



\section{Lexicographic Optimisation of CVaR}

In this section we present the contributions of this paper. We begin with a formal problem statement.
\begin{problem}
	\label{prob:lexicographic_cvar}
	Let $\mathcal{M}$ be an MDP.
	Find the policy $\pi^*$ that optimises the expected cost subject to the constraint that $\pi^*$ obtains the optimal CVaR at confidence level $\alpha$:
	\begin{align}
	\begin{split}
		\pi^*= & \argmin_{\pi \in \Pi^\mathcal{M}_\mathcal{H}}\  \mathbb{E} \big[\mathcal{C}_{\pi}^\mathcal{M}\big], \\
		&\textrm{ s.t.} \hspace{10pt}  \textnormal{CVaR}_\alpha (\mathcal{C}_\pi^\mathcal{M}) = \min_{\pi' \in \Pi_\mathcal{H}^\mathcal{M}} \textnormal{CVaR}_\alpha (\mathcal{C}^\mathcal{M}_{\pi'}).
		\end{split}
	\end{align} 
\end{problem}
\vspace{-0.5mm}
Our approach to this problem extends the approach to CVaR optimisation from~\citet{chow2015risk}, which we have outlined. We begin by emphasising that in the approach in~\citet{chow2015risk}, some histories in the original MDP have zero probability under the adversarial perturbations obtained by solving Eq.~\ref{eq:prop1}. For example, all histories passing through $s_1$ in Fig.~\ref{fig:example_mdp} are not reachable under the adversarial perturbations, as illustrated  in Fig.~\ref{fig:example_mc}. Intuitively, this suggests that such histories do not contribute to the CVaR, as the CVaR can be computed as the expected value under the perturbed transition probabilities~(Proposition~\ref{prop:chow}). 

In this paper, we investigate finding an alternative policy to execute when we reach such histories which are assigned zero probability by the adversary in the CVaR SSPG. By finding an appropriate alternative policy to execute in these situations, we are able to optimise the expected value whilst maintaining the optimal CVaR.
We now state two  properties of these histories which will be useful to ensure the policies obtained by our algorithm maintain the optimal CVaR. 
Full proofs are in the supplementary material.

\begin{proposition}
	\label{prop:2}
	 Let $\pi$ denote a CVaR-optimal policy for confidence level $\alpha$, and let $\sigma$ denote the corresponding adversary policy from Eq.~\ref{eq:prop1}.
%
%
For some history, $h$, if $\mathcal{P}^\mathcal{M}_\pi(h) > 0$ and $\mathcal{P}^\mathcal{G}_{(\pi, \sigma)} (h) = 0$, there exists a policy $\pi' \in \Pi^\mathcal{M}_\mathcal{H}$ which may be executed from $h$ onwards for which the total cost received over the run is guaranteed to be less than or equal to  VaR$_\alpha(\mathcal{C}_\pi^\mathcal{M})$.
\end{proposition}

\noindent \textit{Proof Sketch}:
We prove Proposition~\ref{prop:2} by showing that the original policy, $\pi$, satisfies this condition. Denote by $\mathcal{H}_{g}^\mathcal{M}$ the set all of histories ending at a goal state, and let $h_g \in \mathcal{H}_{g}^\mathcal{M}$ denote any such history. Using the CVaR representation theorem in Equation~\ref{eq:repr_theorem}, we can show that if $h_g$ has non-zero probability under $\pi$ in the original MDP, but is assigned zero probability by the adversary in the CVaR SSPG, then the total cost of $h_g$ must be less than or equal to VaR$_\alpha(\mathcal{C}_\pi^\mathcal{M})$:
\begin{multline}
	\label{eq:proof_eq_in_paper}
	  \mathcal{P}^\mathcal{M}_{\pi} (h_g) > 0 \textnormal{ and }  \mathcal{P}^\mathcal{G}_{(\pi, \sigma)} (h_g) = 0 \implies \\ \cumul(h_g) \leq \textnormal{VaR}_\alpha(\mathcal{C}^\mathcal{M}_\pi).
\end{multline}
Now consider any history, $h$, which has not reached the goal. Assume that $\mathcal{P}^\mathcal{M}_{\pi} (h) > 0$ and $\mathcal{P}^\mathcal{G}_{(\pi, \sigma)} (h) = 0$. For all histories $h_g \in \mathcal{H}_{g}^\mathcal{M}$ reachable after $h$ under $\pi$ (i.e. for which $\mathcal{P}^\mathcal{M}_{\pi} (h_g) > 0$), we have that $\mathcal{P}^\mathcal{G}_{(\pi, \sigma)} (h_g) = 0$. Therefore, by Equation~\ref{eq:proof_eq_in_paper} all histories $h_g$ reachable after $h$ under $\pi$ are guaranteed to have $\cumul(h_g) \leq \textnormal{VaR}_\alpha(\mathcal{C}^\mathcal{M}_\pi)$. 
This proves that by continuing to execute $\pi$, the total cost received over the run is guaranteed to be less than or equal to  VaR$_\alpha(\mathcal{C}_\pi^\mathcal{M})$.

\smallskip

Proposition~\ref{prop:2} shows that if a state is reached which is assigned zero probability by the adversary, then there must exist a policy which can be executed from that state onwards for which the total cost is guaranteed not to exceed the VaR. The following proposition states that by switching to any policy that guarantees that the cost will not exceed the VaR, the optimal CVaR is maintained.

\begin{proposition}
	\label{prop:3}
	During execution of policy $\pi$ optimal for $\textnormal{CVaR}_\alpha (\mathcal{C}^\mathcal{M}_\pi)$, we may switch to any policy $\pi'$ and still attain the same CVaR, provided that $\pi'$ is guaranteed to incur total cost of less than or equal to $\textnormal{VaR}_\alpha (\mathcal{C}^\mathcal{M}_\pi)$. 
\end{proposition}

\noindent \textit{Proof Sketch}: $\textnormal{CVaR}_\alpha (\mathcal{C}^\mathcal{M}_\pi)$ is computed by integrating over the distribution of costs greater than or equal to $\textnormal{VaR}_\alpha (\mathcal{C}^\mathcal{M}_\pi)$ (Equation \ref{eq:cvar_def}). 
Because switching to $\pi'$ does not result in any costs greater than $\textnormal{VaR}_\alpha (\mathcal{C}^\mathcal{M}_\pi)$, the strategy of switching to $\pi'$ cannot increase the CVaR.
Switching to $\pi'$ cannot decrease the CVaR as $\pi$ already attains the optimal CVaR. 
Therefore, the strategy of switching to an appropriate $\pi'$ must attain the same CVaR.

\subsection{CVaR-Expected-Value}
 
Proposition~\ref{prop:3} establishes that during execution of $\pi_{\textnormal{CV}}$ we can switch to another policy, $\pi'$, without influencing the CVaR, provided that the total cost of executing $\pi'$  is guaranteed not to exceed VaR$_\alpha(\mathcal{C}^\mathcal{M}_{\pi_{\textnormal{CV}}})$. Proposition~\ref{prop:2} establishes that such a policy exists if the current history is assigned zero probability by the adversary in the CVaR SSPG. However, as we shall illustrate in the following example there may be multiple policies which satisfy this criterion. Of these policies, we would like to find the one which optimises our secondary objective of expected value as stated in Problem~\ref{prob:lexicographic_cvar}. 
 
\begin{example}
Consider again the model in Fig.~\ref{fig:example}. We established in Example~\ref{example:1} that the optimal CVaR at $\alpha = 0.1$ is 10, which can be computed as the expected value under the Markov chain in Fig.~\ref{fig:example_mc}. The corresponding VaR at $\alpha = 0.1$ is also 10. All of the histories passing through $s_1$ are assigned zero probability by the adversarial perturbations in the CVaR SSPG. At $s_1$, we can continue to execute any policy for which all histories are guaranteed to reach the goal with total cost less than or equal to the VaR threshold of 10. Executing either $d$ or $e$ at $s_1$ satisfies this property, and maintains the optimal CVaR of 10. The approach of~\citet{chow2015risk}, which we refer to as \textit{CVaR-Worst-Case}, would choose action $e$ as in this situation it optimises for the minimum worst-case cost (see Remark~\ref{remark:1}). However, $d$ achieves better expected value and still attains the optimal CVaR of 10. Therefore, to solve Problem~\ref{prob:lexicographic_cvar}, the policy should choose $d$. On the other hand, executing $c$ attains a sub-optimal CVaR of greater than 10, as some histories would receive a total cost of 20.
\end{example}
 
We wish to develop a general approach to finding the policy, $\pi'$, which optimises the expected value, subject to the constraint that the worst-case total cost must not exceed VaR$_\alpha(\mathcal{C}^\mathcal{M}_{\pi_{\textnormal{CV}}})$. We know that such a policy exists for histories assigned zero probability by the adversary in the CVaR SSPG. Therefore, by switching from $\pi_{\textnormal{CV}}$ to $\pi'$ when such histories occur, we can optimise the expected value while still attaining the optimal CVaR, thus solving Problem~\ref{prob:lexicographic_cvar}. 

We first compute the minimum worst-case total cost at each state, $V_{worst}(s)$, using the following minimax Bellman equation which assumes that the agent always transitions deterministically to the worst-case successor state:
\vspace{-2mm}
\begin{multline}
	\label{eq:best_worst_case}
	V_{worst}(s) = \\ \min_{a \in A} \Big[C(s, a) +   \max_{s' \in S}  \Big( \mathbbm{1} (T(s, a, s') > 0) \cdot V_{worst}(s') \Big) \Big],
\end{multline}
\noindent where $\mathbbm{1}$ denotes the indicator function. Let $\pi_{worst}$ denote the policy corresponding to $V_{worst}$. We also compute the optimal worst-case $Q$-values:
 \begin{equation}
 	Q_{worst}(s, a) = C(s, a) +  \max_{s' \in S}  \mathbbm{1}(T(s, a, s') > 0) \cdot V_{worst}(s').
 \end{equation}

Now, assume that the history so far is $h$, and that the cost so far in the history is $\cumul(h)$. To maintain the optimal CVaR according to Proposition~\ref{prop:3}, we can allow an action $a$ to be executed at $h$ if
\begin{equation}
	\label{eq:action_constr}
	\cumul(h) + Q^*_{worst}(s, a) \leq \textnormal{VaR}_\alpha(\mathcal{C}^\mathcal{M}_{\pi_{\textnormal{CV}}}).
\end{equation}
 For any action $a$ which satisfies Eq.~\ref{eq:action_constr}, after taking $a$ we can then execute $\pi_{worst}$ and guarantee that the total cost will not exceed VaR$_\alpha(\mathcal{C}^\mathcal{M}_{\pi_{\textnormal{CV}}})$. Thus, by finding a policy which optimises the expected value subject to the constraint that actions must satisfy Eq.~\ref{eq:action_constr}, we can find the policy with best expected value which is guaranteed to have total cost less than or equal to VaR$_\alpha(\mathcal{C}^\mathcal{M}_{\pi_{\textnormal{CV}}})$ and maintain the optimal CVaR.

Note that which actions are allowed in Eq.~\ref{eq:action_constr} depends not only on the current state, but also on the cost received so far. To arrive at an offline solution we create an augmented MDP, $\mathcal{M}' = (S', A, T', C', G' ,s_0')$, where we restrict which actions can be executed depending on the cost received so far. We start by augmenting the state space of the original MDP  such that  $S' = S \times [0, \textnormal{VaR}_\alpha(\mathcal{C}^\mathcal{M}_{\pi_{\textnormal{CV}}})]$.
This augmented state keeps track of how much cost has been incurred. The action set $A$ is the same as the original MDP.
The transition function ensures that actions are only enabled if they satisfy Eq.~\ref{eq:action_constr}, and propagates the accumulated cost appropriately:
\begin{align}
\label{eq:trans_aug}
\begin{split}
	&T'((s, c), a, (s', c')) = \\
	&\left\{ \begin{array}{ll}
T(s, a, s') & \textrm{if } c + Q_{worst}(s, a) \leq \textnormal{VaR}_\alpha(\mathcal{C}^\mathcal{M}_{\pi_{\textnormal{CV}}}) \\
	\ &  \textrm{and } c' = c + C(s, a),\\
	0 & \textrm{otherwise.}
	 \end{array}
	 \right.
	 \end{split}
\end{align}

The cost function is $C'((s, c), a) = C(s, a)$; the goal states are $G' = G \times [0, \textnormal{VaR}_\alpha(\mathcal{C}^\mathcal{M}_{\pi_{\textnormal{CV}}})]$; and the initial state is $s'_0=(s_0, 0)$. The value function corresponding to the following standard MDP Bellman equation is the optimal expected value subject to the constraint that $\textnormal{VaR}_\alpha(\mathcal{C}^\mathcal{M}_{\pi_{\textnormal{CV}}})$ will not be exceeded for any possible history.
\vspace{-2mm}
\begin{multline}
	\label{eq:augmented_value}
	V^{\mathcal{M}'}(s, c) = \\ 
	\ \ \ \ \min_{a \in A} \Bigg[C'(s, a) + \sum_{(s', c')} T' \big((s, c), a, (s', c') \big) \cdot V^{\mathcal{M}'}(s', c') \Bigg].
\end{multline}
\vspace{-3mm}

Solving Eq.~\ref{eq:augmented_value} is not straightforward due to the additional continuous state variable which prohibits standard discrete value iteration. Therefore, we instead apply value iteration with linear interpolation~\cite{bertsekas2008approximate} for the continuous variable in the same manner as~\cite{chow2015risk}.






We write $\pi'$ to denote the optimal policy corresponding to the value function $V^{\mathcal{M}'}$. By first executing $\pi_{\textnormal{CV}}$, and then executing $\pi'$ on histories assigned zero probability by the adversary in the CVaR SSPG, we switch to using the policy with best expected value subject to the constraint that the optimal CVaR is maintained. Therefore, this approach solves Problem~\ref{prob:lexicographic_cvar}. This approach, which we refer to as \textit{CVaR-Expected-Value} is described in Algorithm~\ref{alg:cvar-expected-value}.

\setlength{\textfloatsep}{10pt}
\begin{algorithm2e}[bt]
	
	\small
	\SetKwIF{If}{ElseIf}{Else}{if}{:}{else if}{else}{end}%
	\SetKwFor{While}{while}{:}{end while}%
	\SetKwFor{ForEach}{foreach}{:}{end foreach}%
	\SetKwProg{function}{function}{}{end}%
	\SetKwRepeat{Do}{do}{while}%
	\DontPrintSemicolon
	
	get $\pi_{\textnormal{CV}}$ and $V_{\textnormal{CV}}(s, y)$ by solving Equation~\ref{eq:cvar_vi} \;
	get $\pi'$ by solving Equation~\ref{eq:augmented_value}\;
	$s^+ \leftarrow s^+_0$\;
	$c \leftarrow 0$ \;
	
	\caption{\textit{CVaR-Expected-Value}}
	\SetKwProg{function}{function}{}{end}
	\SetKwFunction{executeEpisode}{ExecuteEpisode}
	\function{\executeEpisode{}}{
		\Do{$\xi(s') > 0$}{
			$a \leftarrow \pi_{\textnormal{CV}}(s^+)$ \;
			$\xi = \argmax_{\xi \in \Xi(s, y, a)} \sum_{s' \in S} \xi(s') \cdot T(s, a, s') \cdot V_{\textnormal{CV}}(s', y\cdot \xi(s'), \bot)$ \;
			$s^+ \leftarrow (s', y \cdot \xi(s'), \bot)$, where $s'$ is successor after executing $a$ in the MDP \;
			$c \leftarrow c + C(s, a)$ \;
			\If{$s^+ \in G^+$}{\KwRet}
		}	
		\;
		\While{$s \notin G$}{
			$a \leftarrow \pi'(s, c)$ \;
			$s \leftarrow s'$, where $s'$ is successor after executing $a$ in the MDP \;
			$c \leftarrow c + C(s, a)$ \;
		}
		
		\KwRet
	}
	\label{alg:cvar-expected-value}
\end{algorithm2e}

In this approach, we decide which actions are pruned out based on VaR$_\alpha(\mathcal{C}^\mathcal{M}_{\pi_{\textnormal{CV}}})$. To estimate VaR$_\alpha(\mathcal{C}^\mathcal{M}_{\pi_{\textnormal{CV}}})$, one approach would be to compute the worst-possible history in the Markov chain induced by Eq.~\ref{eq:prop1}. However, this is computationally challenging as the number of reachable histories in the continuous state space may be very large. Therefore, we take the simple approach of first executing $\pi_{\textnormal{CV}}$ (\textit{CVaR-Worst-Case}) for many episodes and compute a Monte Carlo estimate for VaR$_\alpha(\mathcal{C}^\mathcal{M}_{\pi_{\textnormal{CV}}})$~\cite{hong2014monte}, which we then use to restrict actions according to Eq.~\ref{eq:trans_aug}.
\section{Experiments}
The code and data used to run the experiments is included in the supplementary material and will be made publicly available. We experimentally evaluate the following three approaches:  \textit{CVaR-Worst-Case} (\emph{CVaR-WC}), \textit{CVaR-Expected-Value} (\emph{CVaR-EV}), and \textit{Expected Value} (\emph{EV}). \textit{Expected Value}   is the policy which optimises the expected value only. All algorithms are implemented in C++ and Gurobi is used to solve linear programs where necessary. We use 30 interpolation points for $\alpha$ to solve Eq.~\ref{eq:cvar_vi}, and 100 interpolation points for the cumulative cost to solve Eq.~\ref{eq:augmented_value}. The experiments used a 3.2 GHz Intel i7 processor with 64 GB of RAM.

We compare the approaches on the following four domains. The three synthetic domains are from the literature, and the fourth domain we introduce is based on real data.

\textbf{Inventory Control (IC)} We consider the stochastic inventory control problem~\cite{puterman2005markov, ahmed2017sampling}, with 10 decision stages. 
The current number of units in the inventory is $n$, and the maximum number of units is $N=20$.
The action set is $a \in \{0, \ldots, N-n\}$, representing the amount of stock to purchase at each stage.  
There is an expense of $p_u=1$ for purchasing each item of stock.
Items are sold according to the stochastic demand for each stage, $d \in \{0, \ldots, N\}$. 
A revenue of $r = 3$ is received for each unit sold.
For any items of stock which do not sell, there is a holding expense at each stage of $p_h = 1$ per unit.
Thus, the profit received at each stage is 
$$ min \{d, n+a\}\cdot r - a \cdot p_u - max \{n + a - d, 0 \} \cdot p_h
$$

We assume that the demand is modelled as random walk. At each stage, $d = d_{prev} + \Delta d$, where $d_{prev}$ is the demand at the previous stage, and $\Delta d$ is uniformly distributed between $\pm 5$. At the first stage, $d_{prev} = 10$. The demand at the previous stage is included in the state space.

We model all domains as SSP MDPs, where a positive cost function is minimised. 
To convert rewards to a positive cost function, we take the standard approach~\cite{kolobov2012planning} of defining the cost function to be the maximum possible reward minus the reward received. %
In Inventory Control, the maximum possible profit is: $\textnormal{max}(\textrm{profit}) = 400$. The cost for an episode is $\textnormal{max}(\textrm{profit})$ minus the cumulative profit received over all stages. Therefore, a cost of under 400 represents a net profit, and a cost over 400 represents a net loss.

\textbf{Betting Game (BG)} We adapt this domain from the literature on CVaR in MDPs~\cite{bauerle2011markov}. The state is represented by two factors: $(money, stage)$. The agent begins with $money~=~5$. The amount of money that the agent can have is limited between $0$, and $\textnormal{max}(money) = 100$. At each stage the agent
may choose to place a bet from $bets = \{0, 1, 2, 3, 4, 5\}$ provided that sufficient money is available. If the agent wins
the game at that stage, an amount of money equal to the bet
placed is received. If the agent loses the stage, the money bet is lost. If the agent wins the jackpot, the agent receives $10\times$ the amount bet. For each stage, the probability of winning is $0.7$, the probability of winning the jackpot is $0.05$, and the probability of losing is $0.25$. After 10 stages, the cost received is $\textnormal{max}(money) - money$.

\textbf{Deep Sea Treasure (DST)}
This domain is adapted from the literature on multi-objective optimisation in MDPs~\cite{vamplew2011empirical}. A submarine navigates a gridworld to collect one of many treasures, each of which is associated with a reward value, $r$ (illustrated in the supplementary material). At each timestep, the agent chooses from 8 actions corresponding to directions of travel and moves to the corresponding square with probability 0.6 and each of the adjacent squares with probability 0.2. The episode ends when the agent reaches a treasure or the horizon of 15 steps is reached. The agent incurs a cost of 5 at each timestep, and a terminal cost of $500 - r$.  There is a tradeoff between risk and expected cost as if the submarine travels further it may collect more valuable treasure, and therefore incur less cost, but risks running out of time before reaching any treasure.

\begin{figure}[t!]
	\centering
	\includegraphics[width=0.85\columnwidth]{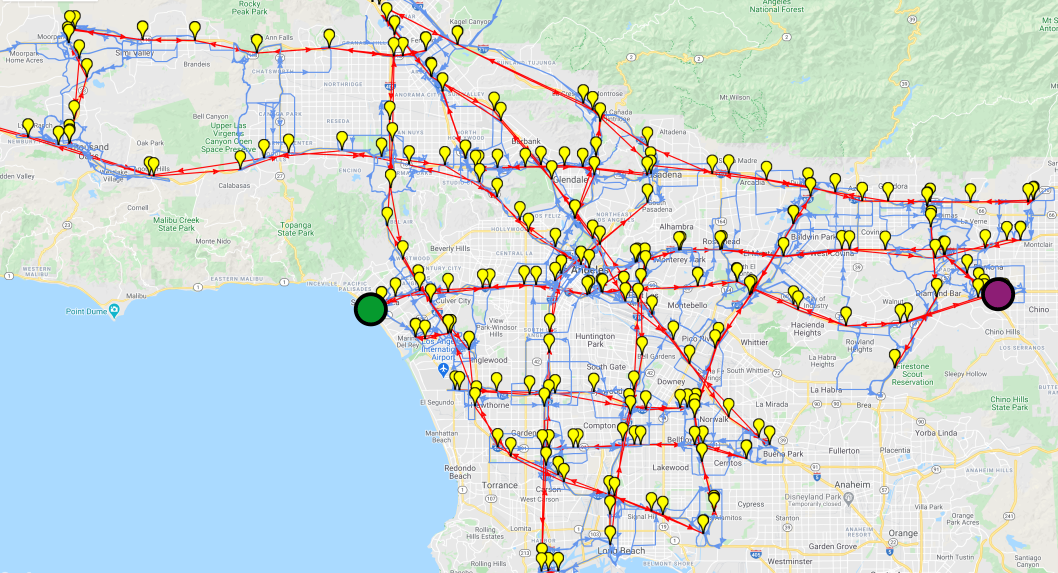}
	\caption{Autonomous Vehicle Navigation domain: MDP state space (yellow balloons) and transitions (red and blue edges) overlaid onto map of Los Angeles, USA. The green and purples circles indicate the start and goal locations. }
	\label{fig:roads}
\end{figure}

\textbf{Autonomous Navigation (AN)} An autonomous vehicle must plan routes across Los Angeles, USA between a start and goal location as illustrated in Fig.~\ref{fig:roads}. 
We access real road traffic data collected by Caltrans Performance Measurement System (PeMS) by over 39,000 real-time traffic sensors deployed across the major metropolitan areas of California~\cite{chen2001freeway}. 
We select a subset of 263 sensors along the major freeways, shown as yellow markers in Figure~\ref{fig:roads}. 
We specify two types of transitions: $i)$~\textit{freeway} transitions (red) along a specified freeway where the transition time distribution is generated from historical PeMS traffic data (discrete with 10 bins), and $ii)$~\textit{between-freeway} transitions (blue) where each state is connected to its three nearest neighbours on other freeways and the transition time is normally distributed around the expected duration obtained from querying the Google Routes API. In this domain, the cost is the journey duration in minutes.
To simulate rare traffic jams on the freeways, uniform noise in [0, 0.1] is added to the slowest \textit{freeway} transitions (and probabilities renormalised). This is motivated by knowledge that rare but severe traffic can affect transition durations on freeways, and introduces a tradeoff between risk and expected cost. 

\begin{table*}[t]
	\small
	\caption{Results from evaluating each method for 20,000 episodes. Rows indicate the optimisation method. Columns indicate the performance for the CVaR and expected value objectives in each domain. The bolded results indicate the method with best performance for expected value given that the CVaR objective is optimal (i.e. Problem~\ref{prob:lexicographic_cvar}). Brackets indicate standard errors. \vspace{-1mm}}
	\renewcommand{\arraystretch}{1.07}
	\begin{subtable}{\columnwidth}
		\resizebox{2.1\columnwidth}{!}{%
			\begin{tabular}{ |@{}l@{\hspace{2mm}} | c@{\hspace{1mm}} c@{\hspace{1mm}} | c@{\hspace{1mm}}  c@{\hspace{1mm}} | c@{\hspace{1mm}}c@{\hspace{1mm}} | c@{\hspace{1mm}}c@{\hspace{1mm}} |}  \cline{2-9}
				\multicolumn{1}{c|}{} & \multicolumn{2}{c|}{Inventory Control (IC)} & \multicolumn{2}{c|}{Betting Game (BG)} & \multicolumn{2}{c|}{Deep Sea Treasure (DST)} & \multicolumn{2}{c |}{Autonomous Navigation (AN)} \\
				\hline
				Method & CVaR$_{0.02}$  & Expected Value & CVaR$_{0.02}$ & Expected Value & CVaR$_{0.02}$ &  Expected Value & CVaR$_{0.02}$ & Expected Value\\ 
				\hline 
				\textit{CVaR-WC} ($\alpha \hspace{0.5mm}\textnormal{=}\hspace{0.5mm} 0.02$) & 386.49 (0.23)  & 286.18 (0.50) & \textbf{95.0} (0.0) & \textbf{95.0} (0.0) & 502.25 (0.78) & 402.74 (0.43) & 210.20 (1.36) & 167.35 (0.11)) \\
				\textit{CVaR-EV} ($\alpha \hspace{0.5mm}\textnormal{=}\hspace{0.5mm} 0.02$) & \textbf{386.92} (0.24)  & \textbf{250.38} (0.66) & \textbf{95.0} (0.0) & \textbf{95.0} (0.0) & \textbf{502.98} (0.83)  & \textbf{352.06} (0.57) & \textbf{211.16} (1.40) & \textbf{164.14} (0.12) \\ 
				\textit{Expected Value} & 416.42 (0.60) & 235.62 (0.70) & 100.0 (0.0) & 58.26 (0.22) & 575.0 (0.0) & 315.15 (0.66) & 315.99 (2.12) & 120.19 (0.38) \\ \hline
				Method & CVaR$_{0.2}$ & Expected Value & CVaR$_{0.2}$ & Expected Value & CVaR$_{0.2}$ & Expected Value & CVaR$_{0.2}$ & Expected Value\\ \hline
				\textit{CVaR-WC} ($\alpha \hspace{0.5mm}\textnormal{=}\hspace{0.5mm} 0.2$) & 360.65 (0.31) & 272.51 (0.48) & 91.97 (0.08) & 82.95 (0.06) & 422.29 (0.62) & 349.45 (0.42) & 172.05 (0.80) & 134.12 (0.21) \\
				\textit{CVaR-EV} ($\alpha \hspace{0.5mm}\textnormal{=}\hspace{0.5mm} 0.2$)  & \textbf{360.29} (0.31) &  \textbf{250.08} (0.63) & \textbf{91.86} (0.08) & \textbf{75.63} (0.16) & \textbf{422.80} (0.63) & \textbf{340.12} (0.49) & \textbf{171.58} (0.80) & \textbf{132.77} (0.22) \\
				\textit{Expected Value}  & 370.54 (0.37) & 235.62 (0.70) & 97.36 (0.07) & 58.26 (0.22) & 453.07 (1.09) & 315.15 (0.66) & 217.86 (0.71) & 120.19 (0.38) \\ 
				\hline
			\end{tabular}%
		}
	\end{subtable}
	\label{table:results}
\end{table*}

\subsection{Results}

The results of our experiments are in Table~\ref{table:results}. The rows in the table indicate the method and the confidence level that the method is set to optimise. The columns indicate the performance of the policy for each objective over 20,000 evaluation episodes in each domain. We measure performance for CVaR$_{0.02}$ (i.e.\ the mean cost of the worst 2\% of runs), CVaR$_{0.2}$ (worst 20\%), and the expected value. We expect the \textit{CVaR-EV} method to match the \textit{CVaR-WC} performance on its CVaR measurement column. We also expect \textit{CVaR-EV} to achieve lower cost than \textit{CVaR-WC} in expectation. 

In the Inventory Control domain, \textit{EV} performs the best for expected value, but the worst for the CVaR objectives. 
For the CVaR$_{0.02}$ objective, \textit{EV} obtains 416, representing an average net loss of 16 on the worst 2\% of runs. 
In contrast, when optimising for CVaR$_{0.02}$ (i.e. $\alpha = 0.02$), both \textit{CVaR-WC} and \textit{CVaR-EV} obtain values of 386 (profit of 14) for the CVaR$_{0.02}$ objective. 
This illustrates that the risk-averse approaches avoid the possibility of losses on poor runs.
Similarly, we see that \textit{CVaR-EV} and \textit{CVaR-WC} equally outperform \textit{EV} at optimising the CVaR$_{0.2}$ objective. 
For both $\alpha = 0.02$ and $\alpha = 0.2$, \textit{CVaR-EV} significantly improves the expected value compared to \textit{CVaR-WC}, meaning that the average profitability is improved while still avoiding risks.

Histograms for the total costs received by \textit{CVaR-WC} ($\alpha = 0.02$), \textit{CVaR-EV} ($\alpha = 0.02$), and \textit{EV} are shown in Figure~\ref{fig:hists} for the Inventory Control domain. Equivalent plots for the other domains are in the supplementary material. We see that while \textit{EV} performs the best in expectation, it incurs the most runs where the cost is above 400, corresponding to net losses. Therefore, \textit{EV} performs worse at CVaR$_{0.02}$. For \textit{CVaR-WC} and \textit{CVaR-EV}, the right tail of the distributions are equivalent, resulting in the same performance for CVaR$_{0.02}$. However, for \textit{CVaR-EV}, the left side of the distribution is spread further left, improving the expected value.

Similarly, in the Betting Game domain  \textit{EV} performs the best for expected value, but worse for the CVaR objectives. For \textit{CVaR-WC} and \textit{CVaR-EV} with $\alpha = 0.02$, the optimal policy is never to bet, and this policy attains the best performance for the CVaR$_{0.02}$ objective. For $\alpha = 0.2$, both \textit{CVaR-WC} and \textit{CVaR-EV} achieve similar performance for CVaR$_{0.2}$. However, \textit{CVaR-EV} attains significantly lower cost in expectation. This occurs because winning the jackpot is usually sufficient to guarantee that the VaR will not be exceeded. In these situations,   \textit{CVaR-WC} stops betting. On the other hand, \textit{CVaR-EV} bets aggressively in these situations, as bets can safely be made without the risk of having a bad run which would influence the CVaR. 

For both the Deep Sea Treasure and Autonomous Navigation domains, we make the same observation that both \textit{CVaR-WC} and \textit{CVaR-EV} achieve the same CVaR performance when optimising each of the CVaR$_{0.02}$ and CVaR$_{0.2}$ objectives. However, $\textit{CVaR-EV}$ obtains better expected value performance. In both domains, we also see that \textit{EV} performs the best in expectation, but less well at the CVaR objectives.



The computation times in Table~\ref{table:times} indicate that for three out of four domains, the computation required for \textit{CVaR-EV} is only a moderate (5\%-30\%) increase over the computation required for  \textit{CVaR-WC}. For Inventory Control, the increase in computation time is more substantial (150\%).


\section{Conclusion}
In this paper, we have presented an approach to optimising the expected value in MDPs subject to the constraint that the CVaR is optimal. Our experimental evaluation on four domains has demonstrated that our approach is able to attain optimal CVaR while improving the expected performance compared to the current state of the art method. In future work, we wish to improve scalability by extending our approach to use labelled real-time dynamic programming~\cite{bonet2003labeled} rather than value iteration over the entire state space.

\FloatBarrier
\begin{table}[b]
	\centering
	\caption{Computation times for each approach in seconds.\vspace{-1mm}}
	\hspace{-3mm}
	\resizebox{0.8\columnwidth}{!}{%
		\begin{tabular}{@{}l@{\hspace{3mm}}c@{\hspace{3mm}}c@{\hspace{3mm}}c@{\hspace{3mm}}c@{\hspace{0mm}}} \hline
			Method & IC & BG & DST & AN \\ 
			\hline 
			\textit{CVaR-Worst-Case} & 19637 & 6215 & 8156 & 18327 \\
			\textit{CVaR-Expected-Value} &  48527 & 6526  & 10653 & 23020 \\
			\textit{Expected Value}  &  8303 & 34.5 & 505& 1876  \\
			\hline
		\end{tabular}%
	}%
	\label{table:times}
\end{table}

\begin{figure}[h!]
	\centering
	\begin{subfigure}[t]{0.42\textwidth}
		\caption{\vspace{1.5mm} \hspace{8mm} \textit{CVaR-Worst-Case} ($\alpha = 0.02$) \vspace{-3.5mm} }
		\includegraphics[width=\linewidth]{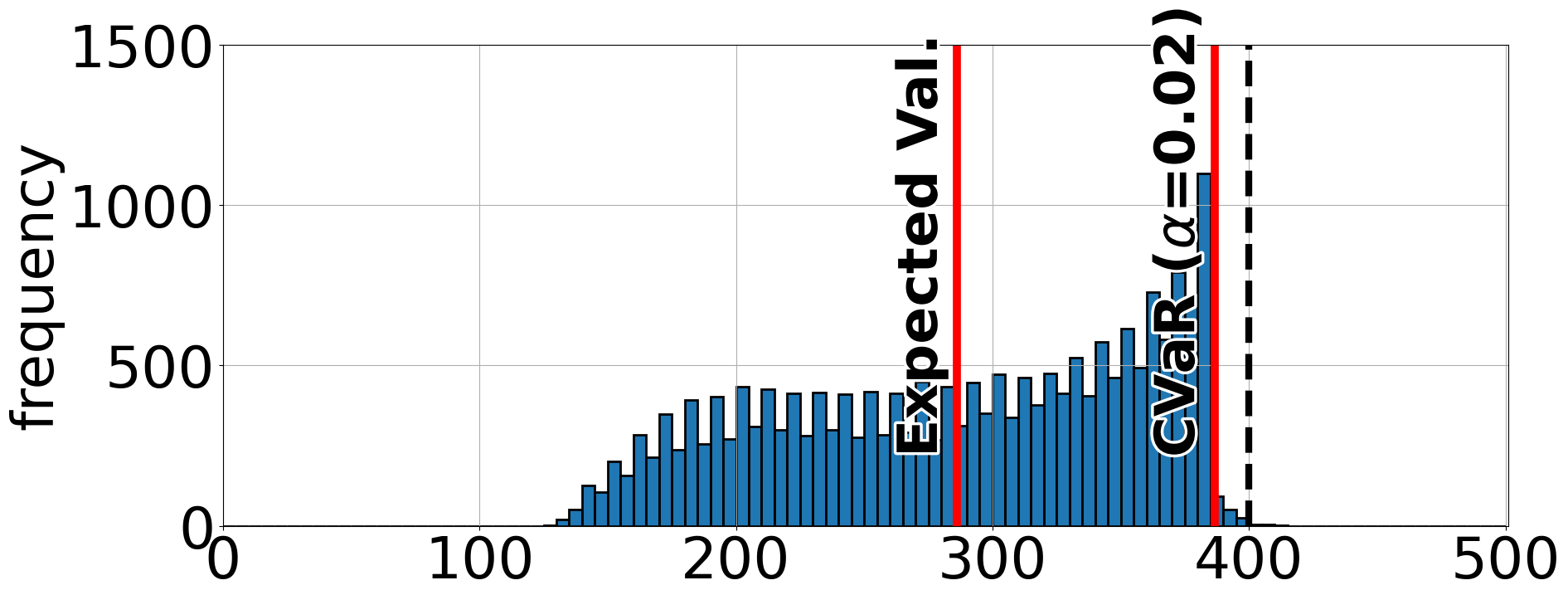}
		\label{fig:cvar-worst-case}
		\vspace{-2mm}
	\end{subfigure}
	
	\begin{subfigure}[t]{0.42\textwidth}
		\caption{\vspace{1.5mm} \hspace{8mm} \textit{CVaR-Expected-Value} ($\alpha = 0.02$) \vspace{-3.5mm} }
		\includegraphics[width=\linewidth]{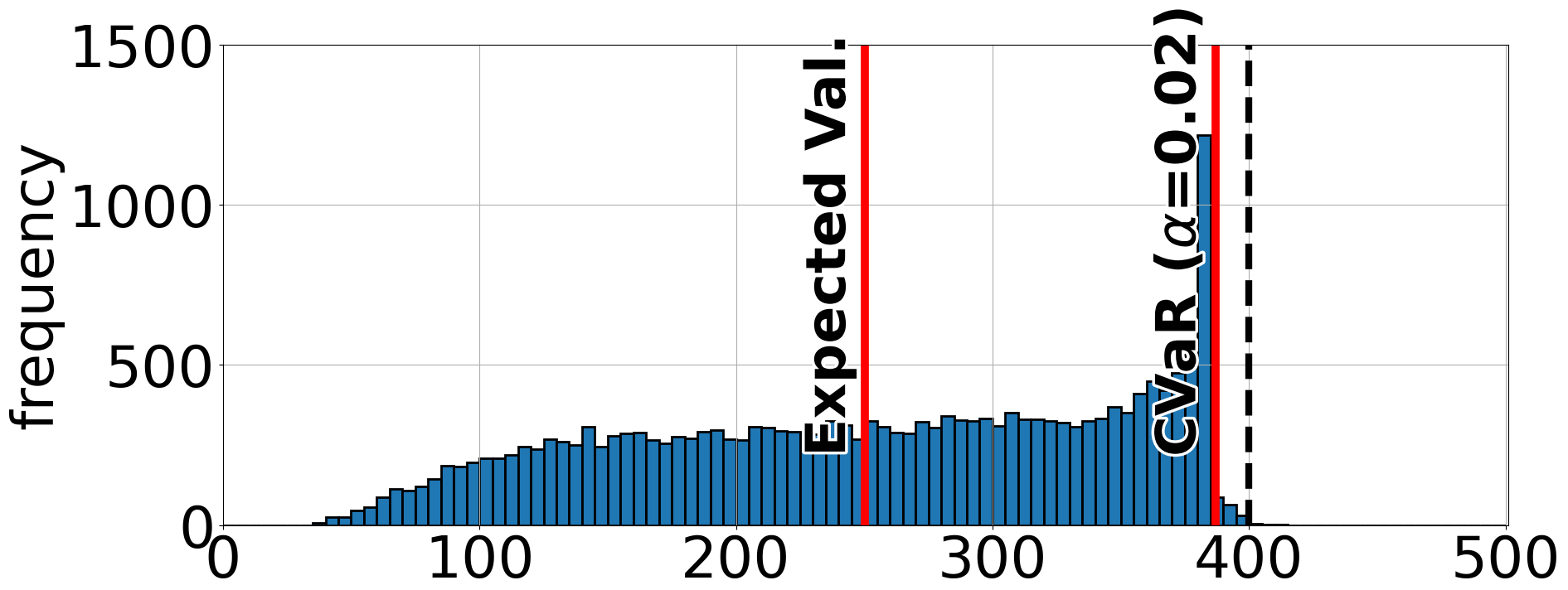}
		\label{fig:cvar-expected}
		\vspace{-2mm}
	\end{subfigure}
	
	\begin{subfigure}[t]{0.42\textwidth}
		\caption{ \vspace{1.5mm} \hspace{8mm} \textit{Expected Value} \vspace{-4mm}}
		\includegraphics[width=\linewidth]{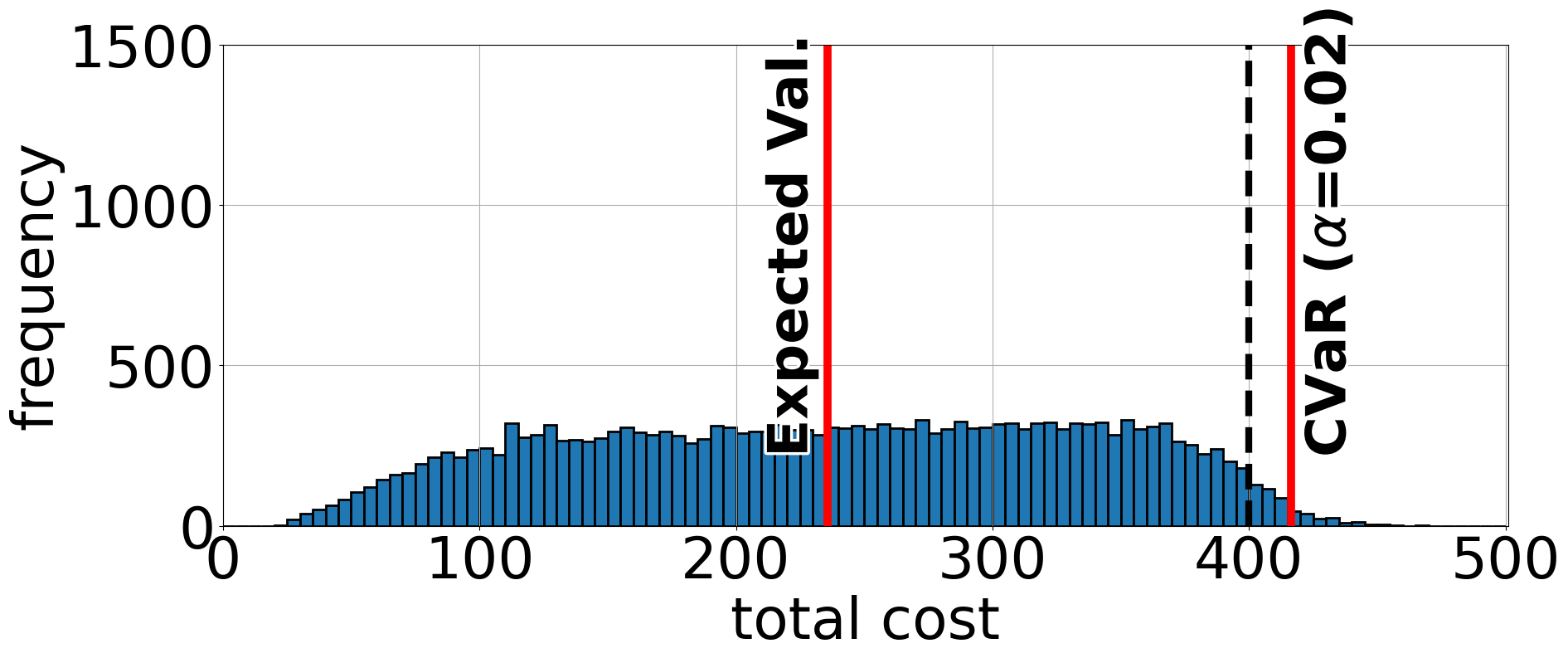}
		\label{fig:expected}
		\vspace{-5mm}
	\end{subfigure}
	
	\caption{Histograms for the total cost received over 20000 evaluation episodes in the Inventory Control domain. Total costs of greater than 400 (dashed black line) represent losses, whereas total costs of less than 400 represent profit.}
	\vspace{-5mm}
	\label{fig:hists}
\end{figure}

\clearpage
\small
\bibliography{references}
\normalsize
\clearpage
\appendix 

\section{Proof of Proposition~\ref{prop:2}}

\textit{Let $\pi$ denote a CVaR-optimal policy for confidence level $\alpha$, and let $\sigma$ denote the corresponding adversary policy from Equation~\ref{eq:prop1}.
	%
	%
	For some history, $h$, if $\mathcal{P}^\mathcal{M}_\pi(h) > 0$ and $\mathcal{P}^\mathcal{G}_{(\pi, \sigma)} (h) = 0$, there exists a policy $\pi' \in \Pi^\mathcal{M}_\mathcal{H}$ which may be executed from $h$ onwards for which the total cost received over the run is guaranteed to be less than or equal to  VaR$_\alpha(\mathcal{C}_\pi^\mathcal{M})$.
}

\begin{proof}
	We prove Proposition~\ref{prop:2} by showing that the original policy $\pi$ must satisfy this condition. Denote by $\mathcal{H}_{g}^\mathcal{M}$ the set all of histories ending at a goal state. We write $h_g \in \mathcal{H}_{g}^\mathcal{M}$ to denote a specific history ending at a goal state. 
	
	Consider a perturbation function, $\delta(h_g)$, such that $0 \leq \delta(h_g)  \leq \frac{1}{\alpha}$ and $\sum_{h_g} \delta(h_g) \cdot \mathcal{P}^\mathcal{M}_\pi(h_g) = 1 $. The expected value of $\mathcal{C}^\mathcal{M}_\pi$ under a distribution perturbed by $\delta$ is:
	\begin{equation}
		\mathbb{E}_\delta[\mathcal{C}^\mathcal{M}_\pi] = \sum_{h_g \in \mathcal{H}_{g}^\mathcal{M}}\delta(h_g) \cdot \mathcal{P}^\mathcal{M}_\pi(h_g) \cdot \cumul(h_g).
	\end{equation} 
	
	By the CVaR representation theorem in Equation~\ref{eq:repr_theorem}, 
	
	\begin{equation}
		\label{eq:mdp_repr_theorem}
		\max_\delta \mathbb{E}_\delta[\mathcal{C}^\mathcal{M}_\pi] = \textnormal{CVaR}_\alpha (\mathcal{C}^\mathcal{M}_\pi).
	\end{equation} 
	
	Let $\mathcal{H}_{g, \textnormal{VaR}}^\mathcal{M} \subseteq \mathcal{H}_{g}^\mathcal{M}$ denote the set of histories for which $\cumul(h_g) = \textnormal{VaR}_\alpha(\mathcal{C}^\mathcal{M}_\pi)$. Any perturbation function which maximises Equation~\ref{eq:mdp_repr_theorem}, denoted by $\delta^*$, must satisfy the following conditions:
	
	
	\vspace{-3mm}
	\begin{align}
		\delta^*(h_g) & = 1/\alpha\hspace{10pt} & \textnormal{if}\hspace{10pt} \cumul(h_g) > \textnormal{VaR}_\alpha(\mathcal{C}^\mathcal{M}_\pi), \label{eq:first_condition}\\
		\delta^*(h_g) & = 0 \hspace{10pt} & \textnormal{if}\hspace{10pt} \cumul(h_g) < \textnormal{VaR}_\alpha(\mathcal{C}^\mathcal{M}_\pi),
	\end{align}

	\vspace{-6mm}
	\begin{equation}
		\label{eq:final_condition}
		\sum_{h_g \in \mathcal{H}_{g, \textnormal{VaR}}^\mathcal{M}} \delta^*(h_g) \cdot \mathcal{P}^\mathcal{M}_\pi(h_g) = 1 - \frac{1}{\alpha}\Pr(\mathcal{C}^\mathcal{M}_\pi > \textnormal{VaR}_\alpha(\mathcal{C}^\mathcal{M}_\pi)),
	\end{equation}
	
	\noindent where Equation~\ref{eq:final_condition} ensures that the perturbed probability distribution sums to 1. Notably, $\delta^*(h_g) = 0$ only if $\cumul(h_g) \leq \textnormal{VaR}_\alpha(\mathcal{C}^\mathcal{M}_\pi)$.
	
	From Proposition~\ref{prop:chow} we also have that:
	\begin{equation}
		\textnormal{CVaR}_\alpha (\mathcal{C}^\mathcal{M}_\pi) = \sum_{h_g \in \mathcal{H}_{g}^\mathcal{M}} \mathcal{P}^\mathcal{M}_{(\pi, \sigma)} (h_g) \cdot  \cumul(h_g).
	\end{equation}
	
	Therefore:
	\begin{align}
		\begin{split}
			\sum_{h_g \in \mathcal{H}_{g}^\mathcal{M}}\delta^*(h_g)   \cdot \mathcal{P}^\mathcal{M}_\pi(h_g) \cdot \cumul(h_g)& = \\
			\sum_{h_g \in \mathcal{H}_{g}^\mathcal{M}} \mathcal{P}^\mathcal{G}_{(\pi, \sigma)} (h_g) \cdot  \cumul(h_g)
			\label{eq:17} & =  \\ 
			\sum_{h_g \in \mathcal{H}_{g}^\mathcal{M}}  \delta_\sigma(h_g)  \cdot \mathcal{P}^\mathcal{M}_\pi(h_g) & \cdot \cumul(h_g)
		\end{split}
	\end{align}
	
	\noindent where we define $\delta_\sigma(h_g) = \mathcal{P}^\mathcal{G}_{(\pi, \sigma)} (h_g)/\mathcal{P}^\mathcal{M}_{\pi} (h_g)$. The values of  $\delta_\sigma(h_g)$ are constrained in the same manner as $\delta(h_g)$ by the definition of the CVaR SSPG. Therefore, Equation~\ref{eq:17} implies that $\delta_\sigma(h_g)$ satisfies the same conditions as $\delta^*(h_g)$ in Equations~\ref{eq:first_condition}-\ref{eq:final_condition}. We have that $\mathcal{P}^\mathcal{G}_{(\pi, \sigma)} (h_g) = 0$ when $\mathcal{P}^\mathcal{M}_{\pi} (h_g) > 0$ only if $\delta_\sigma(h_g) = 0$. $\delta_\sigma(h_g) = 0$ only if $\cumul(h_g) \leq \textnormal{VaR}_\alpha(\mathcal{C}^\mathcal{M}_\pi)$ by  Equations~\ref{eq:first_condition}-\ref{eq:final_condition}. Therefore 
	\begin{multline}
		\label{eq:final_proof_eq}
		\mathcal{P}^\mathcal{G}_{(\pi, \sigma)} (h_g) = 0 \textnormal{ and } \mathcal{P}^\mathcal{M}_{\pi} (h_g) > 0  \implies \\ \cumul(h_g) \leq \textnormal{VaR}_\alpha(\mathcal{C}^\mathcal{M}_\pi).
	\end{multline}
	
	Consider some history, $h$, where $\mathcal{P}^\mathcal{G}_{(\pi, \sigma)} (h) = 0$ and $\mathcal{P}^\mathcal{M}_{\pi} (h) > 0$. For all histories $h_g \in \mathcal{H}_{g}^\mathcal{M}$ reachable after $h$ under $\pi$ (i.e. for which $\mathcal{P}^\mathcal{M}_{\pi} (h_g) > 0$), $\mathcal{P}^\mathcal{G}_{(\pi, \sigma)} (h_g) = 0$. Therefore, by Equation~\ref{eq:final_proof_eq} all histories $h_g$ reachable after $h$ under $\pi$ are guaranteed to have $\cumul(h_g) \leq \textnormal{VaR}_\alpha(\mathcal{C}^\mathcal{M}_\pi)$. This completes the proof that for any history $h$, where $\mathcal{P}^\mathcal{G}_{(\pi, \sigma)} (h) = 0$ and $\mathcal{P}^\mathcal{M}_{\pi} (h) > 0$, by continuing to execute policy $\pi$, the total cost is guaranteed to be less than or equal to $\textnormal{VaR}_\alpha(\mathcal{C}^\mathcal{M}_\pi)$.
\end{proof}

\section{Proof of Proposition~\ref{prop:3}}
\textit{During execution of policy $\pi$ optimal for $\textnormal{CVaR}_\alpha (\mathcal{C}^\mathcal{M}_\pi)$, we may switch to any policy $\pi'$ and still attain the same CVaR, provided that $\pi'$ is guaranteed to incur total cost of less than or equal to $\textnormal{VaR}_\alpha (\mathcal{C}^\mathcal{M}_\pi)$. 
}

\begin{proof}
	Let $\mathcal{H}_{g, \leq \textnormal{VaR}}^\mathcal{M}$ denote the set of histories where $\cumul(h_g) \leq \textnormal{VaR}_\alpha (\mathcal{C}^\mathcal{M}_\pi)$, and $\mathcal{H}_{g, > \textnormal{VaR}}^\mathcal{M}$ denote the set of histories where $\cumul(h_g) > \textnormal{VaR}_\alpha (\mathcal{C}^\mathcal{M}_\pi)$. If for any $h_g \in \mathcal{H}_{g, > \textnormal{VaR}}^\mathcal{M}$, there exists a policy $\pi' \neq \pi$ such that executing $\pi'$ partway through $h_g$ is guaranteed to receive total cost less than or equal to $\textnormal{VaR}_\alpha (\mathcal{C}^\mathcal{M}_\pi)$, then switching to $\pi'$ would improve the CVaR, contradicting that $\pi$ is a CVaR-optimal policy. Therefore, there will never be a suitable $\pi' \neq \pi$ to switch to along any $h_g \in \mathcal{H}_{g, > \textnormal{VaR}}^\mathcal{M}$. 
	
	If partway through any $h_g \in \mathcal{H}_{g, \leq \textnormal{VaR}}^\mathcal{M}$, we switch to a valid $\pi'$ instead of $\pi$, this can result in no total costs above the VaR by the definition of $\pi'$ in Proposition~\ref{prop:3}. Therefore, switching to $\pi'$ cannot modify the distribution of total costs which exceed $\textnormal{VaR}_\alpha (\mathcal{C}^\mathcal{M}_\pi)$.  Additionally, because no total costs which are less than or equal to $\textnormal{VaR}_\alpha (\mathcal{C}^\mathcal{M}_\pi)$ can be increased above $\textnormal{VaR}_\alpha (\mathcal{C}^\mathcal{M}_\pi)$ by switching to $\pi'$, the VaR cannot increase.

	Because switching from $\pi$ to $\pi'$ cannot change the distribution of total costs above the VaR, and it cannot increase the VaR, we observe from the definition of CVaR in Equation~\ref{eq:cvar_def} that the CVaR cannot be increased. Because $\pi$ already attains the optimal CVaR, the strategy of switching to $\pi'$ must attain the optimal CVaR.
\end{proof}

\newpage

\section{Histograms of Total Costs}
In this section, we present histograms showing the total cost received over a number of runs for each method. 

\begin{figure}[h!b]
	\centering
	\begin{subfigure}[t]{0.02\textwidth}
		\vspace{10mm}
		(a)
	\end{subfigure}
	\begin{subfigure}[t]{0.42\textwidth}
		\includegraphics[width=\linewidth, valign=t]{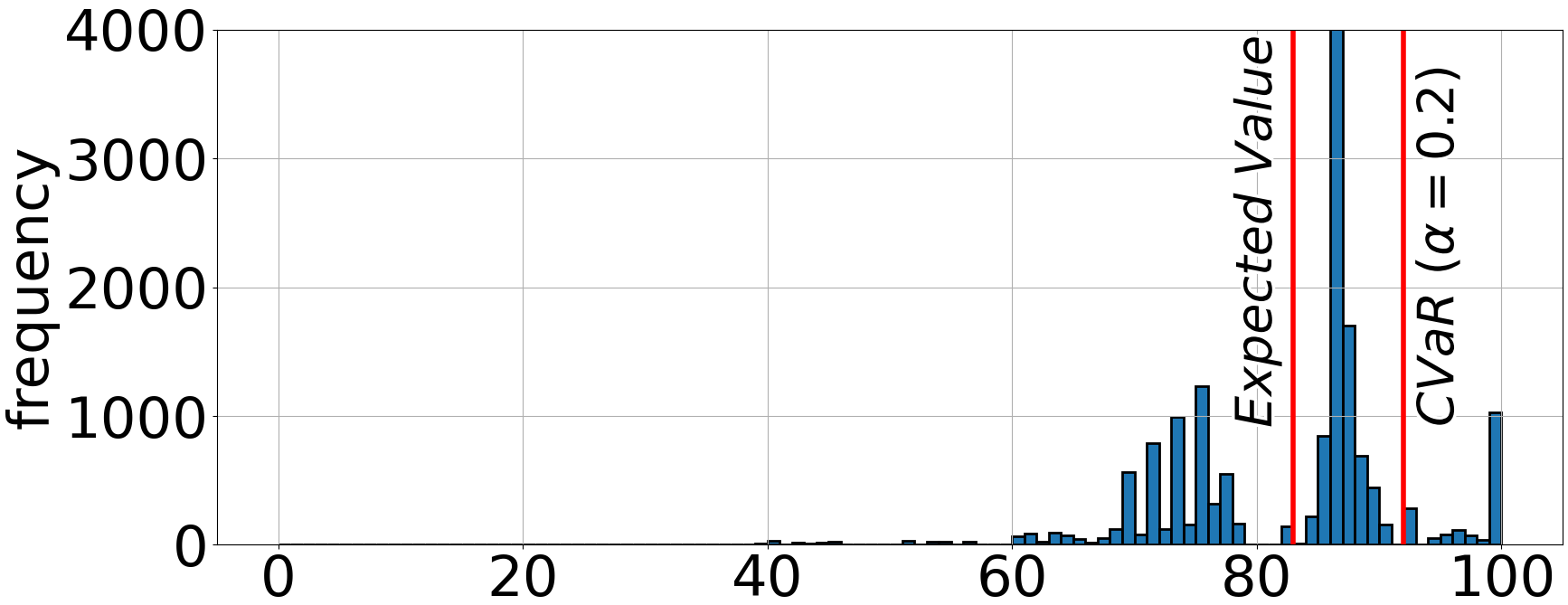}
		\caption{\vspace{1.5mm} \hspace{8mm} \textit{CVaR-Worst-Case} ($\alpha = 0.2$)}
		\label{fig:cvar-worst-case}
		\vspace{-2mm} 
	\end{subfigure}
	
	\begin{subfigure}[t]{0.02\textwidth}
		\vspace{10mm}
		(b)
	\end{subfigure}
	\begin{subfigure}[t]{0.42\textwidth}
		\includegraphics[width=\linewidth, valign=t]{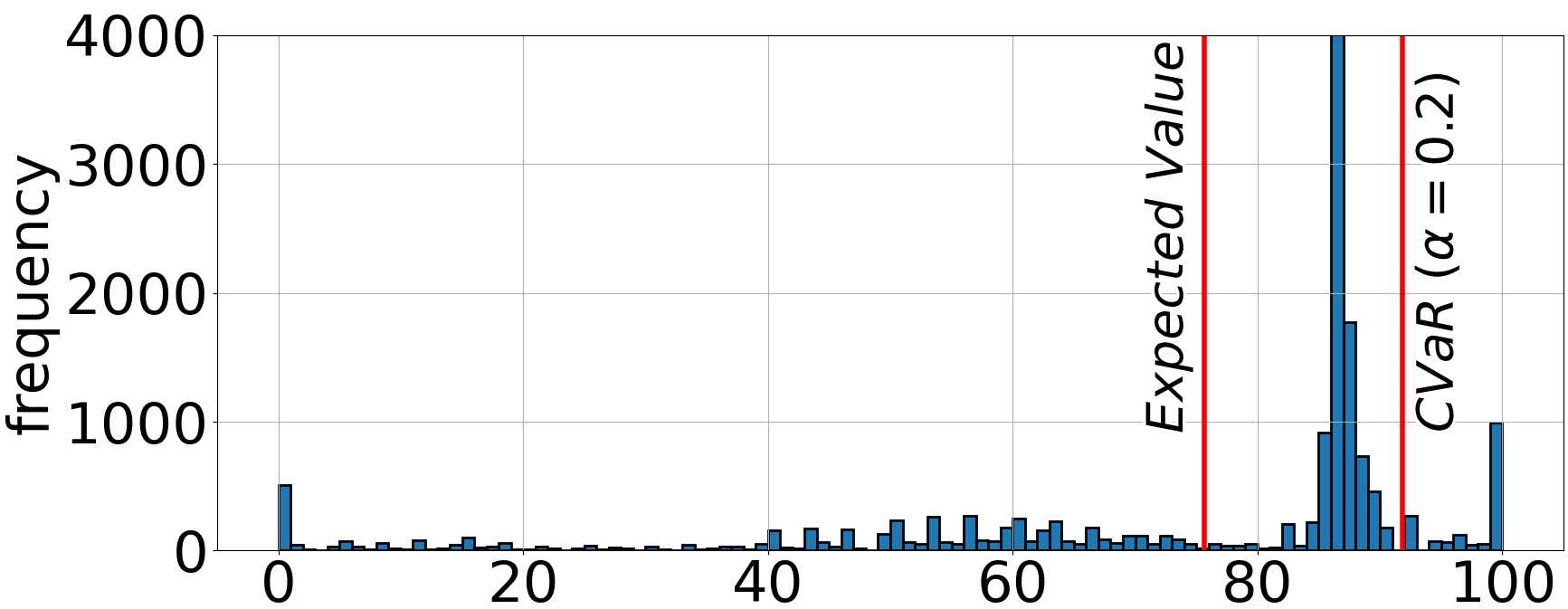}
		\caption{\vspace{1.5mm} \hspace{8mm} \textit{CVaR-Expected-Value} ($\alpha = 0.2$)}
		\label{fig:cvar-expected}
		\vspace{-2mm}
	\end{subfigure}
	
	\begin{subfigure}[t]{0.02\textwidth}
		\vspace{10mm}
		(c)
	\end{subfigure}
	\begin{subfigure}[t]{0.42\textwidth}
		\includegraphics[width=\linewidth, valign=t]{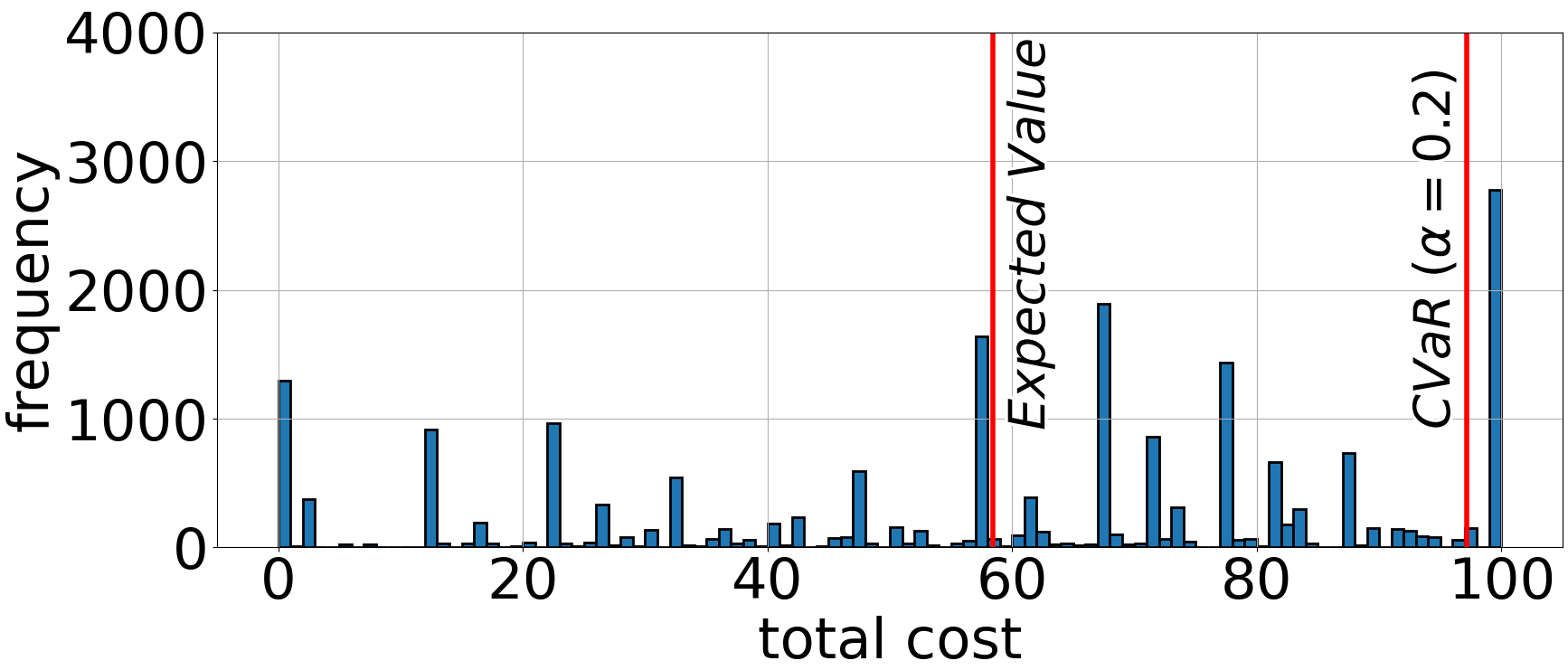}
		\caption{ \vspace{1.5mm} \hspace{8mm} \textit{Expected Value}}
		\label{fig:expected}
		\vspace{-5mm}
	\end{subfigure}
	
	\caption{Histograms for the total cost received over 20000 evaluation episodes in the Betting Game domain.  \label{fig:hist_betting} }
\end{figure}

In Figure~\ref{fig:hist_betting} we present the histograms for the Betting Game domain. We see that for $\alpha = 0.2$, both \textit{CVaR-EV} and \textit{CVaR-WC} have equivalent right tails of the distribution, resulting in equal CVaR$_{0.2}$. However, \textit{CVaR-EV} has more of the distribution shifted towards the left, resulting in better expected cost. \textit{EV} performs well in expectation, but there are many runs which obtain the maximum cost, resulting in poor CVaR performance. 

For the Betting Game domain, we do not include histograms for optimising CVaR $\alpha = 0.02$, as the optimal policy is never to bet and the cost received is therefore always 95 (see Table~\ref{table:results}).

\begin{figure}[t]
	\centering
	\begin{subfigure}[b]{0.42\textwidth}
		\includegraphics[width=1.0\linewidth]{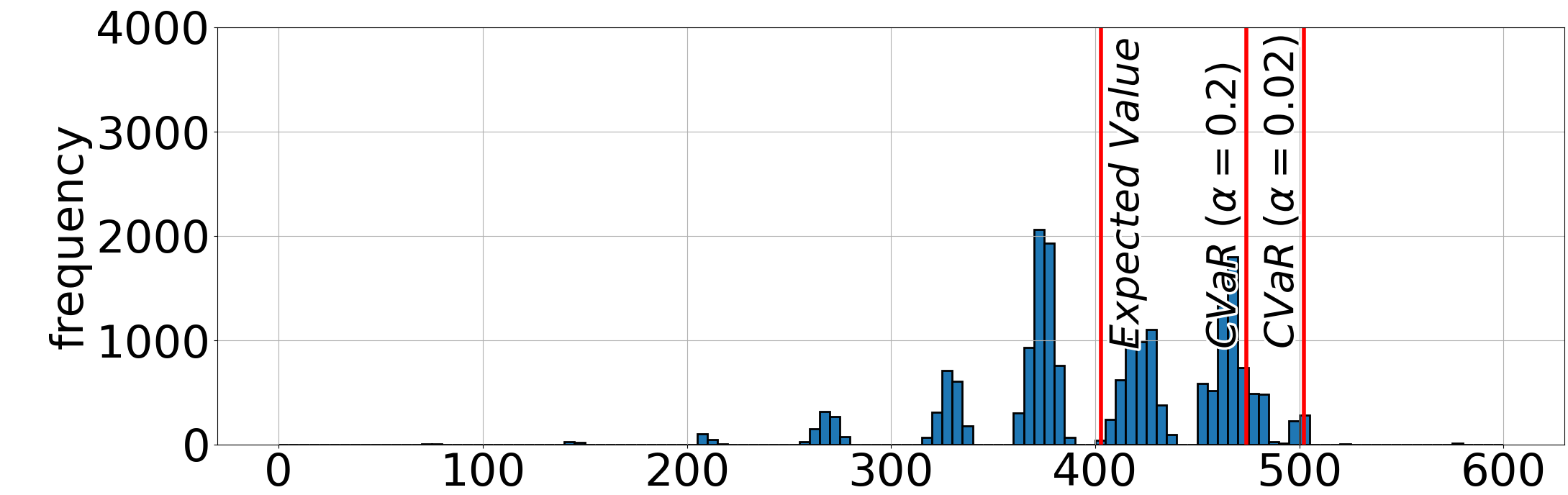}
		\caption{\hspace{10mm} \vspace{-1mm} \textit{CVaR-Worst-Case} ($\alpha = 0.02$)}
	\end{subfigure}
	\vspace{-2mm}
	\begin{subfigure}[b]{0.42\textwidth}
		\includegraphics[width=1.0\linewidth]{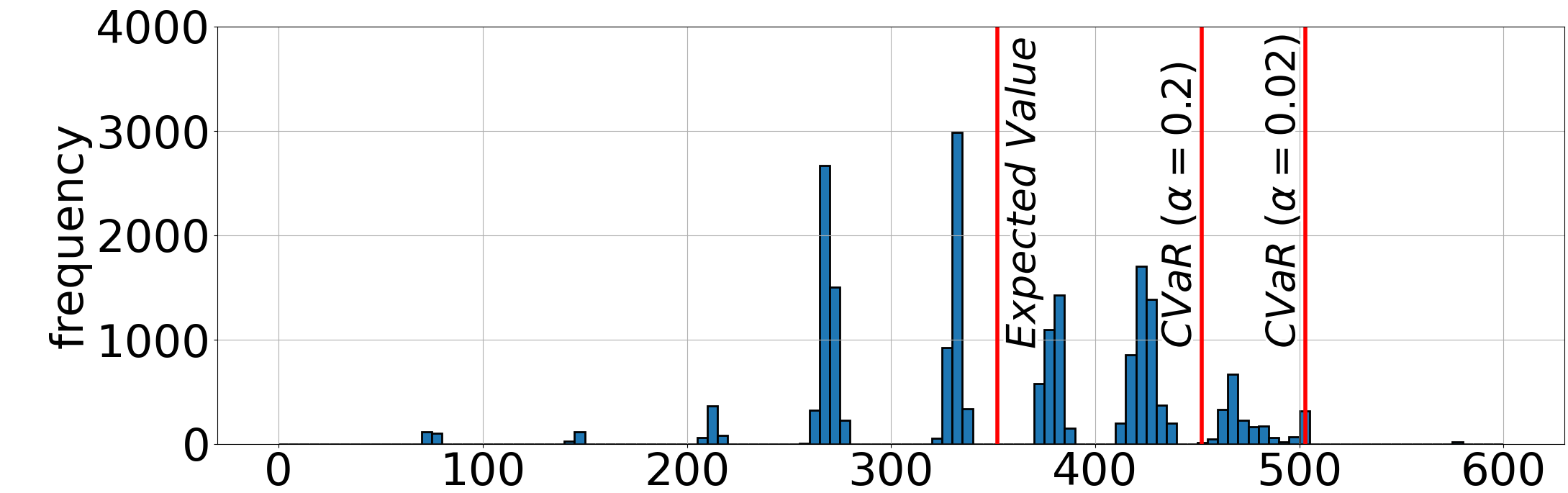}
		\caption{\hspace{10mm} \vspace{-1mm} \textit{CVaR-Expected-Value} ($\alpha = 0.02$)}
	\end{subfigure}
	\vspace{-0mm}
	\begin{subfigure}[b]{0.42\textwidth}
		\includegraphics[width=1.0\linewidth]{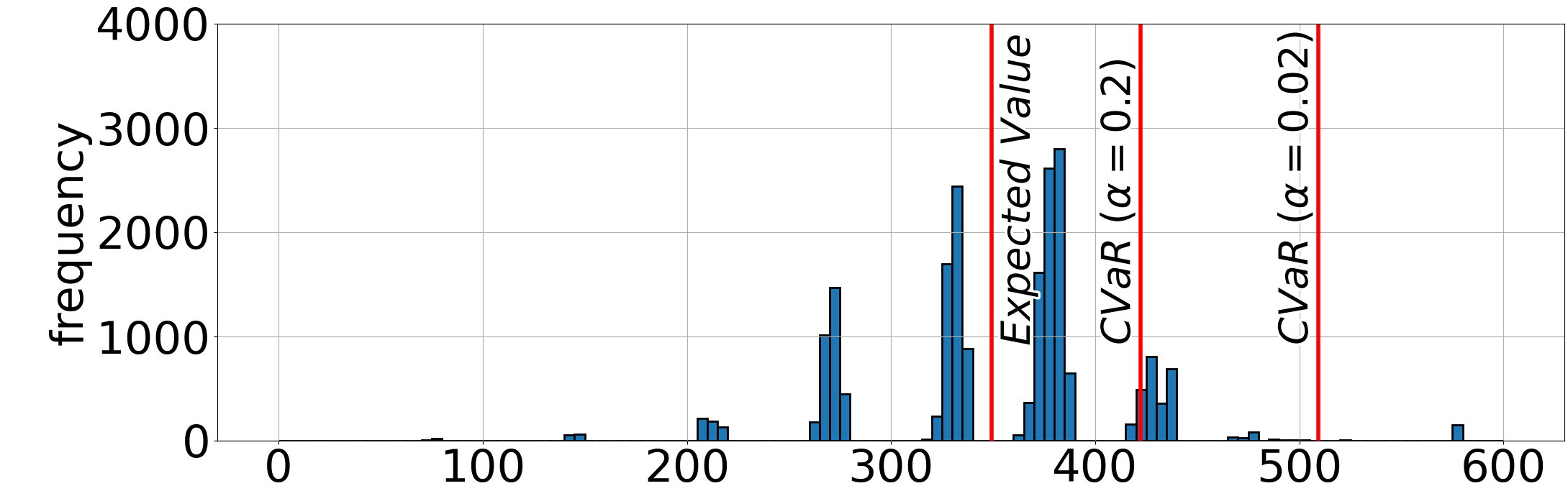}
		\caption{\hspace{10mm} \vspace{-1mm} \textit{CVaR-Worst-Case} ($\alpha = 0.2$)}
	\end{subfigure}
	\vspace{-1mm}
	\begin{subfigure}[b]{0.42\textwidth}
		\includegraphics[width=1.0\linewidth]{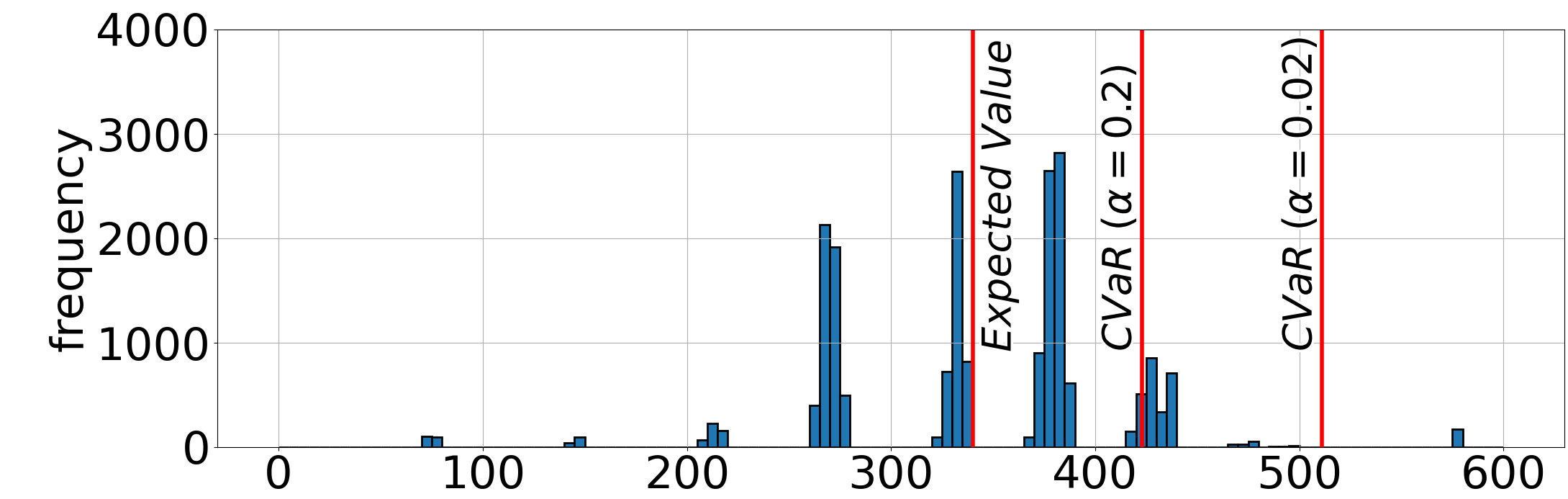}
		\caption{\hspace{10mm} \vspace{-1mm} \textit{CVaR-Expected-Value} ($\alpha = 0.2$)}
	\end{subfigure}
	\vspace{-1mm}
	\begin{subfigure}[b]{0.42\textwidth}
		\includegraphics[width=1.0\linewidth]{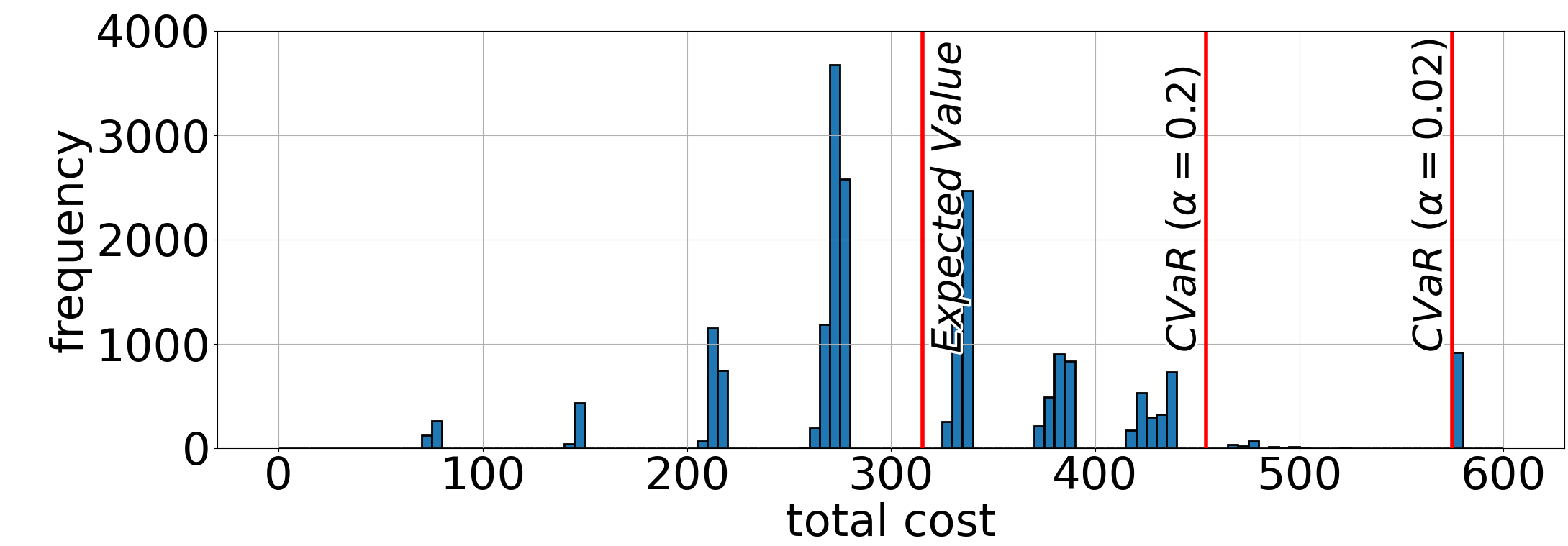}
		\caption{\hspace{10mm} \vspace{-1mm} \textit{Expected Value}}
	\end{subfigure}
	\vspace{-2mm}
	\caption{Histograms for the total cost received over 20000 evaluation episodes in the Deep Sea Treasure domain. \label{fig:hist_dst}}
	\vspace{-3mm}
\end{figure}

Figure~\ref{fig:hist_dst} shows the cost distributions for the Deep Sea Treasure domain. For $\alpha = 0.02$, we see that both \textit{CVaR-EV} and \textit{CVaR-WC} have equivalent distributions in the right most 2\% of the distribution, resulting in equal CVaR$_{0.02}$. However, \textit{CVaR-EV} has the total cost of more runs shifted to the left, resulting in lower expected cost. The same observations are made when optimising with $\alpha = 0.2$, but now the right most 20\% of the distributions are equivalent, resulting in equal CVaR$_{0.2}$. For \textit{EV}, a significant proportion of the runs obtain the maximum possible cost, resulting in poor CVaR performance. 

\begin{figure}[th]
	\centering
	\begin{subfigure}[b]{0.42\textwidth}
		\includegraphics[width=1.0\linewidth]{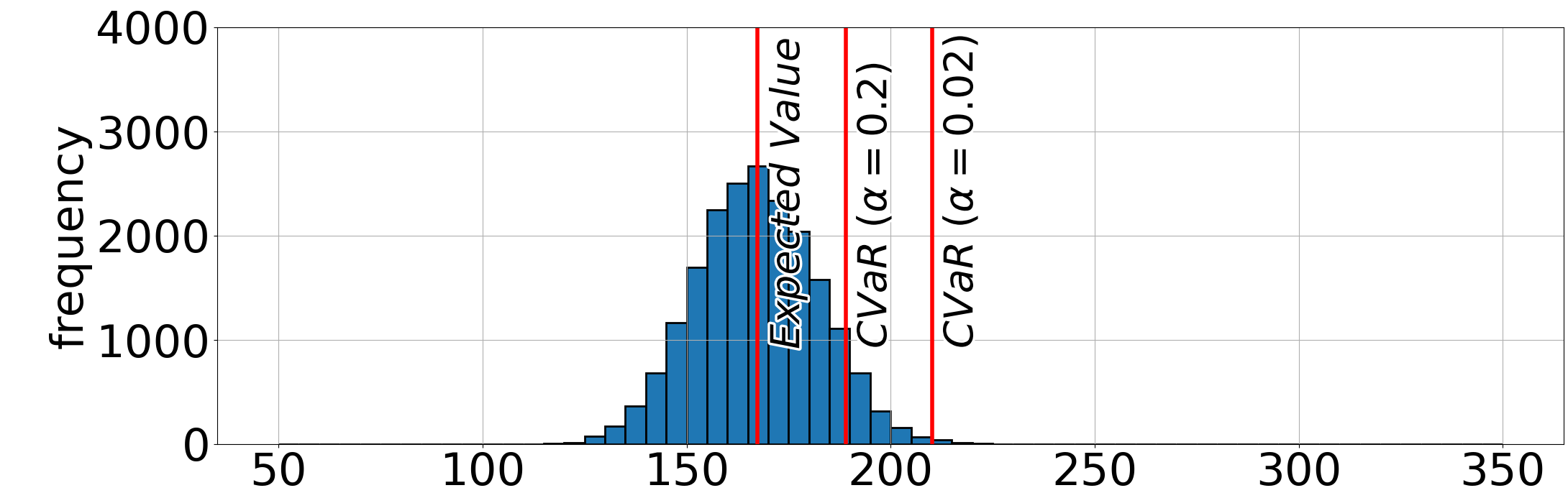}
		\caption{\hspace{10mm} \vspace{-1mm} \textit{CVaR-Worst-Case} ($\alpha = 0.02$)}
	\end{subfigure}
	\vspace{-1mm}
	\begin{subfigure}[b]{0.42\textwidth}
		\includegraphics[width=1.0\linewidth]{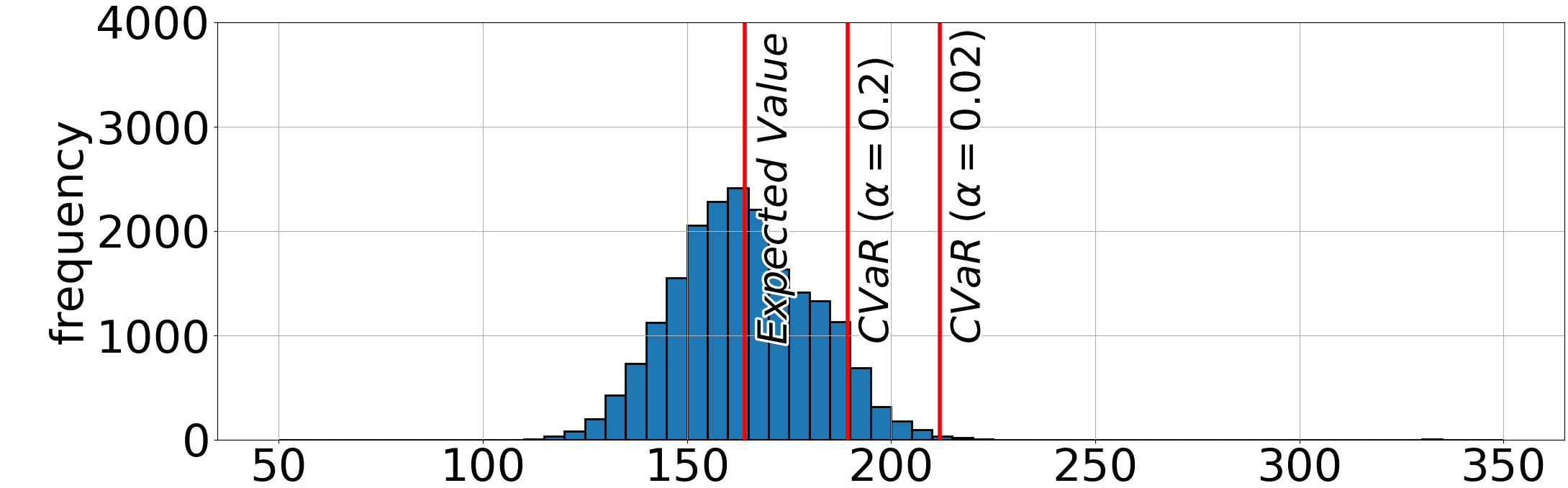}
		\caption{\hspace{10mm} \vspace{-1mm} \textit{CVaR-Expected-Value} ($\alpha = 0.02$)}
	\end{subfigure}
	\vspace{-1mm}
	\begin{subfigure}[b]{0.42\textwidth}
		\includegraphics[width=1.0\linewidth]{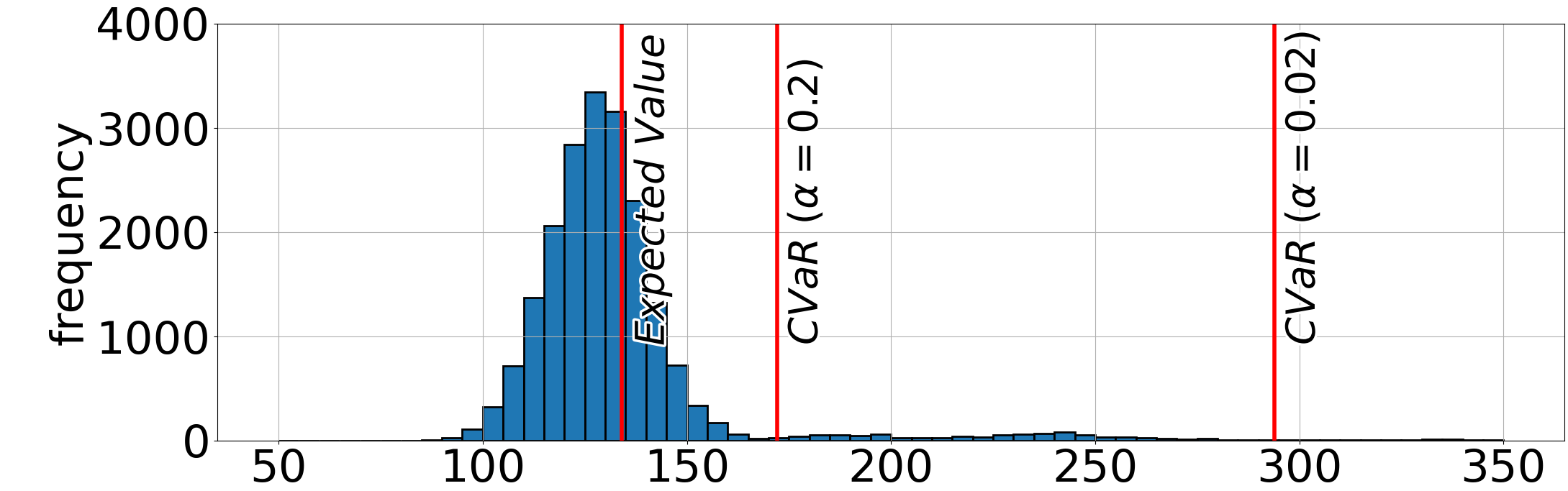}
		\caption{\hspace{10mm} \vspace{-1mm} \textit{CVaR-Worst-Case} ($\alpha = 0.2$)}
	\end{subfigure}
	\vspace{-1mm}
	\begin{subfigure}[b]{0.42\textwidth}
		\includegraphics[width=1.0\linewidth]{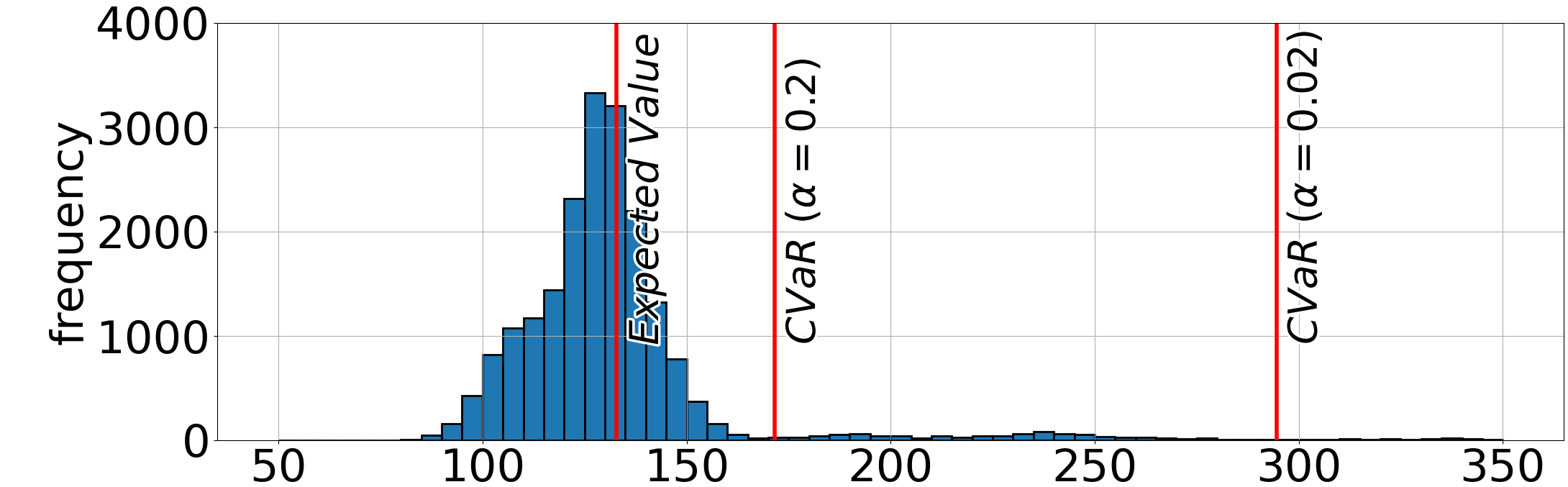}
		\caption{\hspace{10mm} \vspace{-1mm} \textit{CVaR-Expected-Value} ($\alpha = 0.2$)}
	\end{subfigure}
	\vspace{-1mm}
	\begin{subfigure}[b]{0.42\textwidth}
		\includegraphics[width=1.0\linewidth]{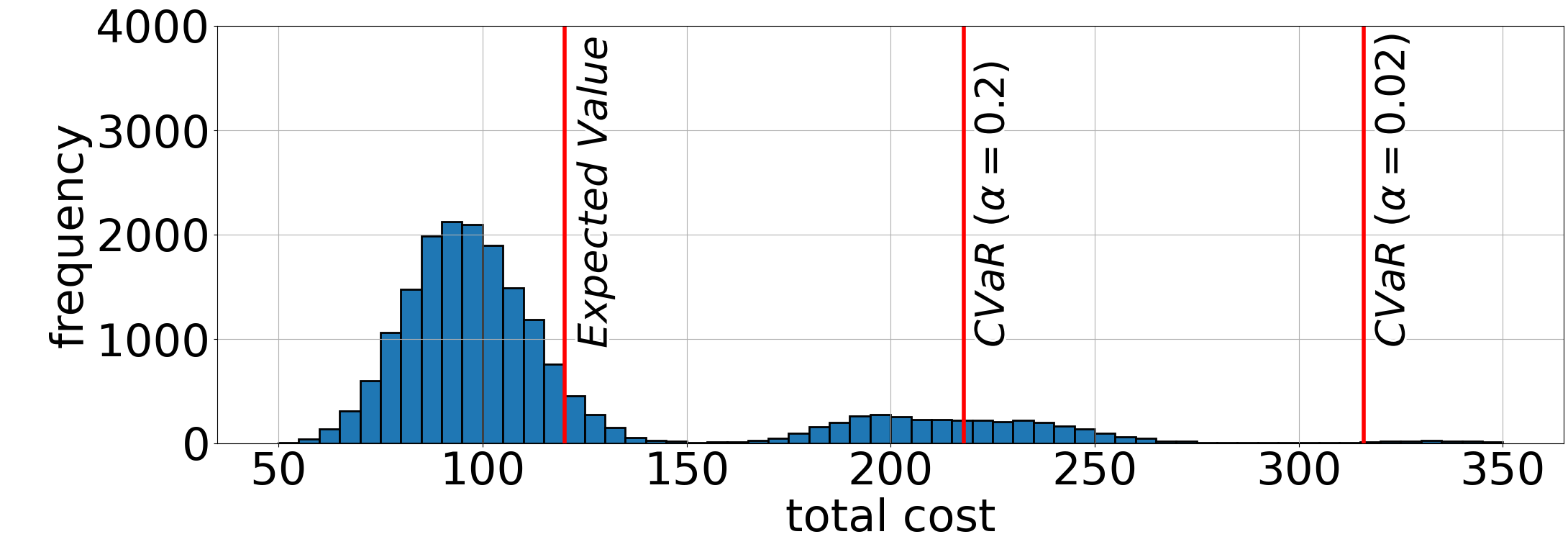}
		\caption{\hspace{10mm} \vspace{-1mm} \textit{Expected Value}}
	\end{subfigure}
	\vspace{-1mm}
	\caption{Histograms for the total cost received over 20000 evaluation episodes in the Autonomous Navigation domain. \label{fig:hist_navigation}}
	\vspace{-3mm}
\end{figure}

The cost distributions for the Autonomous Navigation domain are plotted in Figure~\ref{fig:hist_navigation}. The same observations can be made for this domain as for the other domains. However, the difference between the distributions for \textit{CVaR-EV} and \textit{CVaR-WC} are more modest in the Autonomous Navigation domain.

\newpage 

\FloatBarrier

\newpage 

\section{Illustration of Deep Sea Treasure Domain}
\begin{figure}[h!]
	\centering
	\includegraphics[width=0.47\columnwidth]{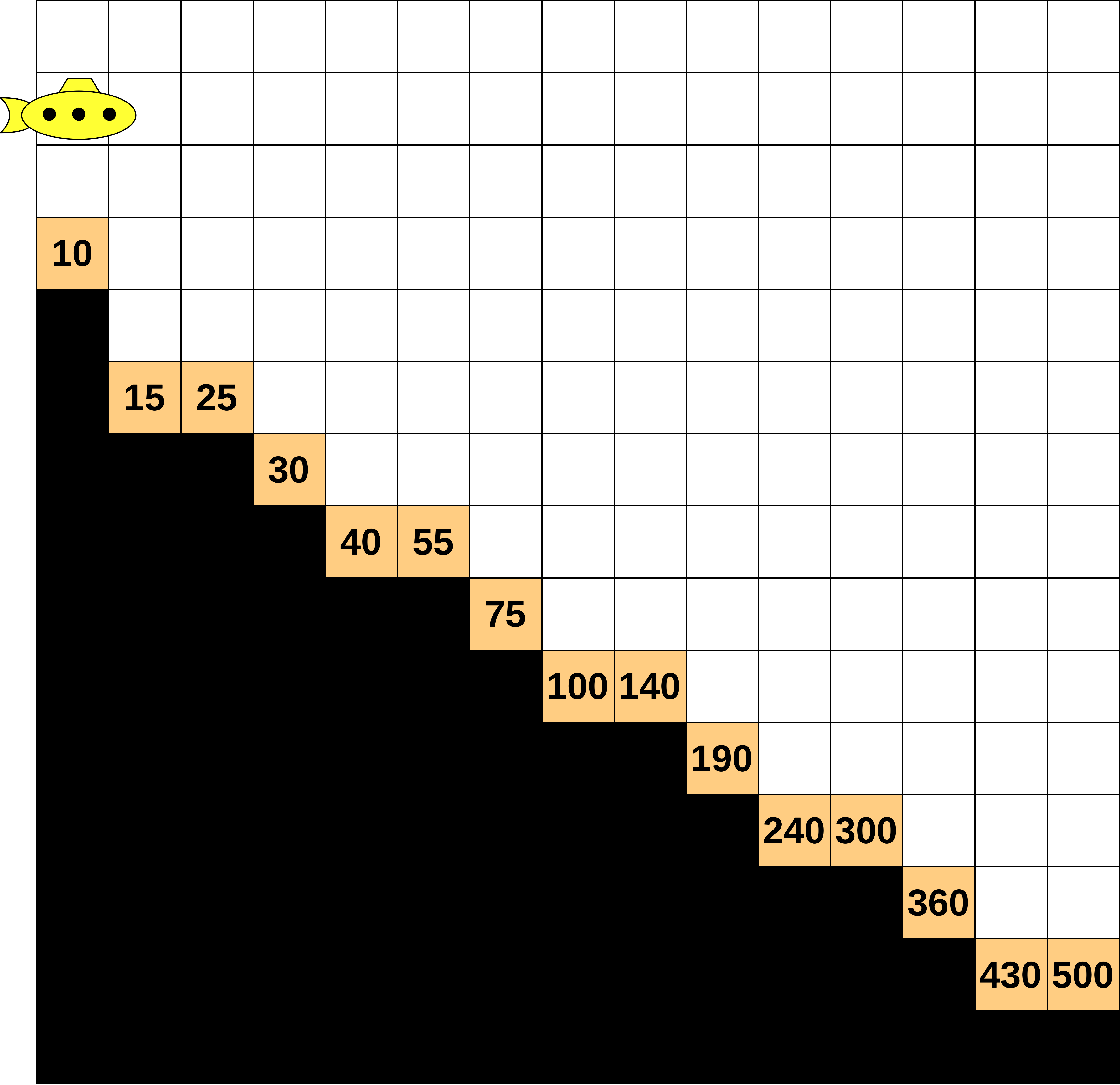}
	\caption{Deep Sea Treasure domain: golden squares indicate different treasure reward values. }
	\label{fig:dst}
\end{figure}

Figure~\ref{fig:dst} illustrates the Deep Sea Treasure domain. 

\end{document}